\documentclass{article} 
\usepackage{iclr2025_conference,times}

\usepackage{amsmath,amsfonts,bm}

\def\eqref#1{equation~\ref{#1}}
\def\Eqref#1{Equation~\ref{#1}}

\def\1{\bm{1}}

\DeclareMathAlphabet{\mathsfit}{\encodingdefault}{\sfdefault}{m}{sl}
\SetMathAlphabet{\mathsfit}{bold}{\encodingdefault}{\sfdefault}{bx}{n}

\usepackage{hyperref}
\hypersetup{hidelinks}
\usepackage{url}
\usepackage{graphicx}
\usepackage{epstopdf}
\usepackage{wrapfig}
\usepackage{algorithm}
\usepackage{algpseudocode}
\usepackage{xcolor}         
\usepackage{caption}
\usepackage{subcaption}
\usepackage{float}
\usepackage{booktabs}
\usepackage{tikz}
\usepackage[toc,page,header]{appendix}
\usepackage{minitoc}

\usepackage{amsthm}
\newtheorem{theorem}{Theorem}
\newtheorem{lemma}{Lemma}

\newtheorem{definition}{Definition}
\newtheorem{assumption}{Assumption}
\newtheorem{proposition}{Proposition}
\numberwithin{lemma}{section}
\numberwithin{theorem}{section}
\numberwithin{corollary}{section}
\numberwithin{definition}{section}
\numberwithin{assumption}{section}
\numberwithin{proposition}{section}
\usepackage{amssymb}

\title{Safe Bayesian Optimization for Complex Control Systems via Additive Gaussian Processes}

\author{Hongxuan Wang \\
National University of Singapore\\
\texttt{hongxuanwang@u.nus.edu} \\
\And
Xiaocong Li \thanks{Corresponding Author.} \\
SIMTech, A*STAR \\
\texttt{li\_xiaocong@simtech.a-star.edu.sg} \\
\And
Lihao Zheng \\
CUHK, Shenzhen \\
\texttt{lihaozheng@link.cuhk.edu.cn} \\
\And
Adrish Bhaumik \& Prahlad Vadakkepat \\
National University of Singapore \\
\texttt{\{adrish07, prahlad\}@nus.edu.sg}
}

\iclrfinalcopy 
\begin{document}


\maketitle

\begin{abstract}
Automatic controller tuning is attractive for robotics and mechatronic systems whose dynamics are difficult to model accurately, but direct black-box optimization can be unsafe because each query is executed on the physical plant. Existing safe Bayesian optimization (BO) methods provide high-probability safety guarantees, yet their practical use in multi-loop control is limited by two coupled difficulties: the controller parameter space is often moderately high-dimensional, and hardware evaluations are too expensive to allow hundreds or thousands of exploratory trials. This paper proposes \textsc{SafeCtrlBO}, a safe BO method for simultaneously tuning multiple coupled controllers. The method uses additive Gaussian-process kernels to encode low-order structure across controller gains and reduce the sample complexity associated with dense full-dimensional kernels. It also replaces the expensive potential-expander computation used in \textsc{SafeOpt}-style exploration with a boundary-based expansion rule that preserves the intended safe-set expansion behavior under explicit geometric conditions and is validated empirically. Experiments on synthetic benchmarks and on a permanent magnet synchronous motor (PMSM) speed-control platform show that \textsc{SafeCtrlBO} reaches high-performing controller parameters with fewer hardware evaluations than representative safe BO baselines, while maintaining the prescribed high-probability safety criterion and avoiding violations of the hard signal-safety constraint in the hardware study. The code implementation is publicly available at \url{https://github.com/hxwangnus/SafeCtrlBO}.
\end{abstract}


\section{Introduction}
\label{sec:intro}

Controller tuning is a recurring bottleneck in robotics and mechatronics. Model-based design gives useful initial controllers, but its final performance depends on the accuracy of simplified plant models, unmodeled friction and saturation effects, delays, and interactions among nested loops. In many deployed systems, engineers therefore still tune gains on hardware after model-based design. This is especially costly for cascade and multi-loop architectures: a field-oriented PMSM controller, for example, contains an outer speed loop and two inner current loops, so six proportional-integral (PI) gains must be selected jointly rather than loop by loop \citep{Gabriel80FOC,Lara16EOR,Wang15FOC}. Similar coupled tuning problems appear in disturbance-observer control \citep{jung2021data}, active-disturbance-rejection control \citep{cao2023improved}, quadrotor control \citep{Berkenkamp16,safe_control-gym}, and precision-motion systems \citep{Rothfuss_gantry3,Wenxin_gantry2,wenxin_gantry1}. A single trial on the physical plant may take minutes and may stress the hardware, so an automatic tuning method must be both sample-efficient and safe.

Data-driven controller optimization avoids identifying a full dynamical model by treating closed-loop performance and safety metrics as black-box functions of the controller parameters. Classical data-driven tuning methods, such as iterative feedback tuning, can be effective when gradient or local sensitivity information is reliable \citep{Hjalmarsson02}, but noisy experiments and nonconvex performance landscapes often make purely local methods brittle. Population-based methods such as genetic algorithms require many trials \citep{Davidor91}. Bayesian optimization (BO) is attractive in this setting because it uses a probabilistic surrogate, typically a Gaussian process (GP), to select informative experiments under a limited evaluation budget \citep{Mockus12,Rasmussen06,Srinivas10}. Standard BO, however, may query controller gains that produce instability, excessive current, or unacceptable tracking errors. Safe BO addresses this by maintaining a certified safe set and evaluating only parameters whose safety constraints are satisfied with high probability \citep{Sui15,Sui18,Berkenkamp16}.

The remaining difficulty is that practical multi-controller tuning sits in an awkward regime for existing safe BO algorithms. The dimension is not extremely high in the machine-learning sense, but it is high enough that dense squared-exponential or Mat\'ern kernels learn slowly from the few hardware trials that are affordable. At the same time, algorithms designed for very high-dimensional safe BO, such as line-search decompositions, may need hundreds or thousands of evaluations before reaching strong performance \citep{Johannes19}, which is often unrealistic on physical control systems. Moreover, \textsc{SafeOpt}-style exploration requires identifying potential expanders of the safe set, a step that becomes computationally expensive as the candidate set and kernel structure grow.

This paper introduces \textsc{SafeCtrlBO} for this moderate-dimensional, hardware-limited regime. The key modeling choice is an additive GP surrogate. Additive kernels represent the objective and safety functions as sums of lower-dimensional main effects and interaction terms \citep{David11,Kirthevasan15,Paul18,Mojmir18,dumbo}. This structure is well matched to controller tuning: different loops often have different gain ranges and time-scale effects, yet the strongest interactions are typically local to subsets of gains. We combine this surrogate with a stagewise safe-optimization procedure. During the expansion stage, we replace the expensive potential-expander search with a boundary-based candidate rule. During the maximization stage, we use a GP-UCB rule restricted to the certified safe set. The resulting method keeps the safety logic of safe BO while reducing the number and computational cost of hardware queries.

\paragraph{Related work.}
\textsc{SafeOpt} was introduced to sequentially optimize unknown functions while avoiding unsafe evaluations \citep{Sui15}. \textsc{StageOpt} separates safe-set expansion from objective maximization and provides convergence guarantees under GP confidence bounds and regularity assumptions \citep{Sui18}. In robotics and control, \textsc{SafeOpt} variants have been used for quadrotor controller tuning \citep{Berkenkamp16}, room-temperature PID tuning \citep{Marcello19}, cascade controller tuning \citep{Khosravi23-SAC}, and other safe learning tasks \citep{turchetta2019safe,bottero2022information}. Recent work has also clarified that practical safety depends critically on valid uncertainty bounds, RKHS-norm assumptions, and discretization choices \citep{fiedler2024safety}. Our theoretical statements therefore make the additional assumptions needed by the boundary-expansion simplification explicit, and our experiments separately report soft performance-threshold violations and hard signal-safety violations.

High-dimensional BO methods reduce sample complexity by exploiting structure. Additive GP models decompose a function into lower-dimensional components and can substantially improve learning efficiency when the true function has additive or low-order interaction structure \citep{David11,Kirthevasan15,Paul18,Mojmir18}. LineBO reduces safe BO to a sequence of one-dimensional subproblems \citep{Johannes19}; it is powerful for high-dimensional search spaces but can require many iterations, which limits its usefulness when each controller evaluation is a hardware experiment. In contrast, \textsc{SafeCtrlBO} targets the common control setting of roughly 6--20 coupled parameters, where a structured full-space surrogate can exploit cross-loop interactions without requiring a long sequence of one-dimensional searches.

\paragraph{Our contributions.}
The contributions are threefold. First, we formulate safe multi-controller tuning with additive GP surrogates, allowing separate signal variances and lengthscales across gain dimensions and interaction orders. Second, we introduce a boundary-based safe-expansion rule that avoids the most expensive potential-expander computation; the paper states the geometric conditions under which this rule is equivalent to searching the outermost expander region and validates the approximation empirically. Third, we evaluate the method on synthetic safe-optimization benchmarks and on a PMSM speed-control hardware platform, showing improved sample efficiency over representative safe BO baselines in the small-evaluation-budget regime relevant to real control experiments.

\begin{figure}[t]
    \centering
    \includegraphics[width=13.9cm]{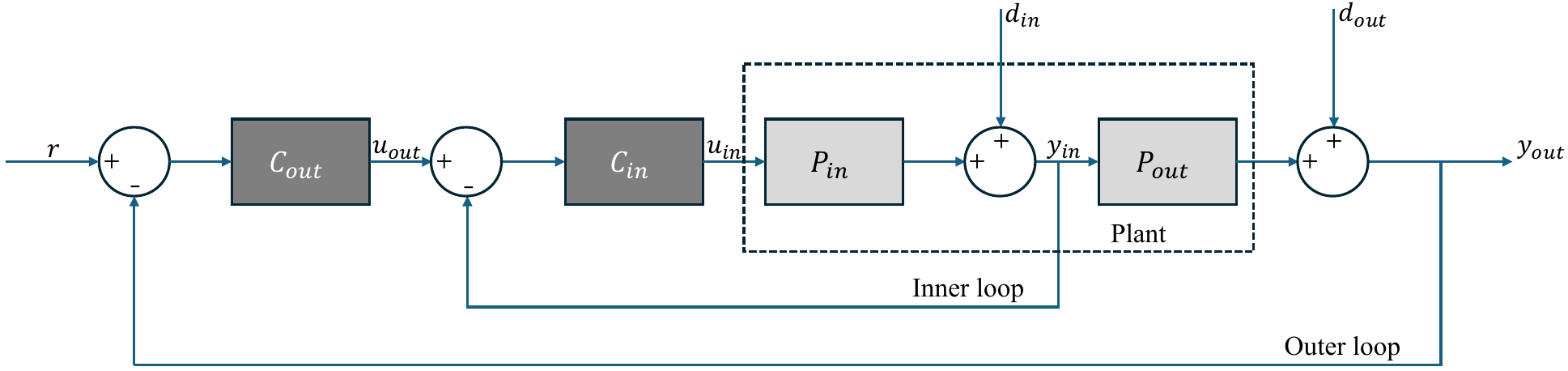}
    \caption{A block diagram for a 2-layer cascade system. The dark grey blocks represent controllers, and the light grey blocks represent plants.}
    \label{fig:cascade}
\end{figure}

\section{Problem Statement}
\label{sec:problem}

We consider automatic tuning for a fixed controller architecture whose parameters are collected in a vector $a\in\mathcal{A}\subset\mathbb{R}^D$. The domain $\mathcal{A}$ is compact and, in the implementation, is represented by a finite candidate set or by a dense discretization from which candidates are sampled. The controller architecture may contain multiple coupled loops. For example, a discrete-time PI controller has
\begin{equation}
    u_k = k_p (r_k-y_k) + k_i \sum_{t=0}^{k}(r_t-y_t),
\end{equation}
where $u_k$ is the control action, $r_k$ is the reference, $y_k$ is the measured output, and $(k_p,k_i)$ are gains. In a two-layer cascade architecture, the outer loop produces the reference for the inner loop:
\begin{equation}
\begin{aligned}
    u_k^{\mathrm{in}} &= k_p^{\mathrm{in}} (u_k^{\mathrm{out}}-y_k^{\mathrm{in}}) + k_i^{\mathrm{in}} \sum_{t=0}^{k}(u_t^{\mathrm{out}}-y_t^{\mathrm{in}}), \\
    u_k^{\mathrm{out}} &= k_p^{\mathrm{out}} (r_k-y_k^{\mathrm{out}}) + k_i^{\mathrm{out}} \sum_{t=0}^{k}(r_t-y_t^{\mathrm{out}}).
\end{aligned}
\end{equation}
More generally, the control signal in layer $\ell$ depends on the reference, the outputs of the outer and inner plants, and the full parameter vector:
\begin{equation}
    u_k^{\ell}=\pi_\ell\!\left(y_k^0,y_k^1,\ldots,y_k^{\ell},r_k,a\right).
\end{equation}
The optimization algorithm does not require an explicit model of these closed-loop dynamics. Instead, it observes scalar performance and safety summaries after executing a trial with controller parameter $a$.

Let $J(a):\mathcal{A}\to\mathbb{R}$ denote the performance objective, chosen so that larger values correspond to better closed-loop behavior. Examples include weighted combinations of settling time, overshoot, steady-state error, or integral error criteria such as ISE, IAE, and ITAE. Let $G_i(a):\mathcal{A}\to\mathbb{R}$, $i=1,\ldots,m$, denote safety or acceptability metrics, each with threshold $h_i$. These metrics may include hard physical limits, such as current, voltage, or input-energy limits, and soft engineering constraints, such as a minimum acceptable performance value or a maximum steady-state error. The safe tuning problem is
\begin{equation}
\label{eq:safe-opt-problem}
\begin{aligned}
    \qquad \qquad \max_{a\in\mathcal{A}} \quad &J(a) \\
    \text{s.t.}\quad &G_i(a)\ge h_i,\quad i=1,\ldots,m.
\end{aligned}
\end{equation}
At iteration $n$, the algorithm selects $a_n$, executes the controller on the plant, and observes noisy measurements
\begin{equation}
\begin{aligned}
    \tilde{J}_n &= J(a_n)+\eta^0_n, \\
    \tilde{G}_{i,n}&=G_i(a_n)+\eta^i_n,
\end{aligned}
\end{equation}
where the noise terms are assumed to be conditionally $R$-sub-Gaussian in the theoretical analysis. We assume at least one initial safe controller $a_0$ is available, together with measurements of $J(a_0)$ and all safety functions $G_i(a_0)$. The goal is to identify a high-performing safe controller using as few hardware evaluations as possible, while ensuring that every evaluated $a_n$ satisfies the safety constraints with high probability.

\section{Safe Bayesian Optimization}
\label{sec:safebo}

Safe BO models the objective and safety functions with GPs and uses confidence intervals to decide which parameters can be evaluated safely. For a generic unknown function $f$ with noisy observations $\tilde{\mathbf{f}}_n=[\tilde f(a_1),\ldots,\tilde f(a_n)]^\top$, the GP posterior mean and variance are
\begin{equation} \label{cal_mean_var}
\begin{aligned}
    \mu_n(a) &= \mathbf{k}_n(a)^\top(\mathbf{K}_n + \sigma_\omega^2\mathbf{I}_n)^{-1} \tilde{\mathbf{f}}_n, \\
    \sigma_n^2(a) &= k(a,a) - \mathbf{k}_n(a)^\top(\mathbf{K}_n + \sigma_\omega^2\mathbf{I}_n)^{-1}\mathbf{k}_n(a),
\end{aligned}
\end{equation}
where $[\mathbf{K}_n]_{ij}=k(a_i,a_j)$ and $\mathbf{k}_n(a)=[k(a,a_1),\ldots,k(a,a_n)]^\top$. We maintain one GP for the objective $J$ and one or many GPs for each safety function $G_i$.

The standard analysis assumes that the true functions have bounded RKHS norms with respect to their kernels and that the observation noise is sub-Gaussian. Under these assumptions, one can choose a confidence width $\beta_n$ such that, with probability at least $1-\delta$,
\begin{equation} \label{eq:confidence interval}
\begin{aligned}
    l_n^{(q)}(a) &\le f^{(q)}(a) \le u_n^{(q)}(a), \\
    u_n^{(q)}(a)&=\mu_{n-1}^{(q)}(a)+\beta_n\sigma_{n-1}^{(q)}(a),\\
    l_n^{(q)}(a)&=\mu_{n-1}^{(q)}(a)-\beta_n\sigma_{n-1}^{(q)}(a),
\end{aligned}
\end{equation}
for all candidate points, functions, and iterations. Here $q=0$ indexes the objective $J$ and $q=1,\ldots,m$ index the safety functions $G_i$. The confidence event in \Eqref{eq:confidence interval} is the source of the high-probability safety statement; in practice it depends on kernel choice, hyperparameter calibration, and the validity of the RKHS and noise assumptions.

The certified safe set at iteration $n$ is
\begin{equation}
\label{eq:safe-set}
    \mathcal{S}_n = \left\{a\in\mathcal{A}\mid l_n^{(i)}(a)\ge h_i,\; i=1,\ldots,m\right\}.
\end{equation}
Safe BO algorithms restrict all evaluations to $\mathcal{S}_n$. \textsc{SafeOpt}-style methods also define potential expanders: safe points whose evaluation could certify new points as safe. In a common notation,
\begin{equation}
\label{eq:expander-set}
    \mathcal{E}_n = \left\{a\in\mathcal{S}_n \mid \exists a'\in\mathcal{A}\setminus \mathcal{S}_n,\; \exists i\in\{1,\ldots,m\}\; \text{s.t.}\; l_{n,(a,u_n^{(i)}(a))}^{(i)}(a')\ge h_i\right\},
\end{equation}
where $l_{n,(a,u_n^{(i)}(a))}^{(i)}$ denotes the lower confidence bound that would be obtained if $G_i(a)$ were observed at its current upper confidence bound. Potential maximizers are safe points that could still be optimal for the objective,
\begin{equation}
    \mathcal{M}_n = \left\{a\in\mathcal{S}_n \mid u_n^{(0)}(a)\ge \max_{a'\in\mathcal{S}_n} l_n^{(0)}(a')\right\}.
\end{equation}
Computing $\mathcal{E}_n$ exactly can be costly because it requires reasoning about the effect of each safe candidate on each currently unsafe candidate. The main algorithmic simplification in \textsc{SafeCtrlBO} is to replace this calculation by a boundary-based approximation during the safe-expansion stage.

\section{SafeCtrlBO}
\label{sec:SafeCtrlBO}

\subsection{Additive Gaussian Kernels}
\label{sec:additive_kernel}

A dense $D$-dimensional squared-exponential kernel treats every dimension as part of a single interaction. This is flexible, but with a small number of hardware trials it often learns slowly in moderate-dimensional controller spaces. \textsc{SafeCtrlBO} instead uses an additive kernel that decomposes the surrogate into lower-dimensional components. For two controller vectors $a,a'\in\mathbb{R}^D$, define the one-dimensional base kernel
\begin{equation}
    z_j(a,a') = \sigma_{f,j}^2 \exp\!\left(-\frac{(a_j-a'_j)^2}{2\ell_j^2}\right), \qquad j=1,\ldots,D,
\end{equation}
where each gain dimension can have its own signal variance $\sigma_{f,j}^2$ and lengthscale $\ell_j$. The order-$q$ additive component is
\begin{equation}
\label{eq:additive-order-q}
    k_{\mathrm{add},q}(a,a') = \sum_{1\le i_1<\cdots<i_q\le D}\; \prod_{r=1}^{q} z_{i_r}(a,a'), \qquad q=1,\ldots,D.
\end{equation}
The full kernel used by the GP can be a weighted sum of selected orders,
\begin{equation}
\label{eq:additive-kernel}
    k_{\mathrm{SafeCtrlBO}}(a,a') = \sum_{q\in\mathcal{Q}} \lambda_q k_{\mathrm{add},q}(a,a'), \qquad \lambda_q\ge 0,
\end{equation}
where $\mathcal{Q}\subseteq\{1,\ldots,D\}$ is chosen by the practitioner. Low orders encode main effects and low-order interactions among controller gains; higher orders can be included when stronger cross-loop interactions are expected. This construction is positive definite because sums and products of positive definite kernels with nonnegative weights are positive definite. Its RKHS and Lipschitz properties are discussed in Appendix~\ref{sec:proof_additive_kernels}.

The modeling advantage is conditional rather than universal. If the objective and safety functions are well approximated by low-order additive interactions, then the effective information gain is governed by the component order instead of the full dimension. For squared-exponential components of maximum order $q_{\max}$, the maximum information gain can be bounded by a sum of component-wise information gains, scaling with the number of included components and the component dimension rather than a single dense $D$-dimensional interaction. If all orders up to $D$ are included, the kernel becomes more expressive but more computationally expensive; in that case the additive structure is still useful as an inductive bias, but the asymptotic dimensional advantage is weaker.

\subsection{Boundary-Based Safe Expansion}
\label{sec:boundary_expansion}

The exact potential-expander set in \Eqref{eq:expander-set} is expensive to compute. We reduce the computational complexity by defining a boundary candidate set. In a continuous domain, the lower-confidence boundary of the safe set is
\begin{equation}
    \mathcal{B}_n = \left\{a\in\mathcal{S}_n \mid \exists i\in\{1,\ldots,m\}: l_n^{(i)}(a)=h_i\right\}.
\end{equation}
On a finite candidate set the equality may be empty, so the implementation uses the relaxed boundary
\begin{equation}
\label{eq:boundary-set}
    \mathcal{B}_n(\tau_n) = \left\{a\in\mathcal{S}_n \mid \min_{i=1,\ldots,m}\left(l_n^{(i)}(a)-h_i\right)\le \tau_n\right\},
\end{equation}
where $\tau_n\ge 0$ is a small tolerance. If $\mathcal{B}_n(\tau_n)$ is empty, we use the safe candidates with the smallest lower-confidence safety margin. This avoids a brittle exact-equality definition and makes the algorithm applicable to sampled candidate sets.

For intuition, let $a_{\mathrm{sb}}\in\mathcal{B}_n$ be a safe-boundary point and let $a_{\mathrm{oes}}(a_{\mathrm{sb}})$ be the closest evaluated safe point. The line segment between them represents the outermost explored region associated with that boundary point:
\begin{equation}
    O_n = \bigcup_{a_{\mathrm{sb}}\in\mathcal{B}_n}\left\{\lambda a_{\mathrm{sb}}+(1-\lambda)a_{\mathrm{oes}}(a_{\mathrm{sb}})\mid \lambda\in[0,1]\right\}.
\end{equation}
When GP variance increases along these outward segments, maximizing uncertainty over the boundary gives the same candidate as maximizing uncertainty over $O_n$. This monotonicity is not a universal property of multi-point GP posteriors; Appendix~\ref{proof_thm3} states a sufficient geometric condition and gives 1D/2D visual checks.

During safe expansion, \textsc{SafeCtrlBO} selects the boundary point with the largest uncertainty among the safety functions,
\begin{equation}
\label{eq:boundary-acquisition}
    a_n = \arg\max_{a\in\mathcal{B}_n(\tau_n)}\; \max_{i=1,\ldots,m}\sigma_{n-1}^{(i)}(a).
\end{equation}
During maximization, it selects the safe point with the largest objective upper confidence bound,
\begin{equation}
\label{eq:max-acquisition}
    a_n = \arg\max_{a\in\mathcal{S}_n} u_n^{(0)}(a).
\end{equation}
The complete procedure is summarized in Algorithm~\ref{algorithm}.

\begin{algorithm}[t]
\caption{\textsc{SafeCtrlBO}}
\label{algorithm}
\begin{algorithmic}[1]
\Statex \textbf{Inputs:} parameter domain/candidate set $\mathcal{A}$; additive kernels for $J$ and $G_i$
\Statex \hspace{12mm} safety thresholds $h_i$, $i=1,\ldots,m$; confidence widths $\beta_n$
\Statex \hspace{12mm} initial safe data $\mathcal{D}_0=\{a_0,\tilde J(a_0),\tilde G_1(a_0),\ldots,\tilde G_m(a_0)\}$
\Statex \hspace{12mm} boundary tolerance sequence $\tau_n$; expansion budget or stopping rule $T_0$
\For{$n = 1,2,\ldots,T_0$}
    \State Fit/update the GP for $J$ and the GPs for all $G_i$ using $\mathcal{D}_{n-1}$
    \State Compute $\mathcal{S}_n \gets \{a\in\mathcal{A}\mid l_n^{(i)}(a)\ge h_i,\; i=1,\ldots,m\}$
    \State Compute $\mathcal{B}_n(\tau_n)$ using \Eqref{eq:boundary-set}
    \State Select $a_n \gets \arg\max_{a\in\mathcal{B}_n(\tau_n)} \max_{i=1,\ldots,m}\sigma_{n-1}^{(i)}(a)$
    \State Evaluate the controller with $a_n$ and observe $\tilde J(a_n),\tilde G_1(a_n),\ldots,\tilde G_m(a_n)$
    \State $\mathcal{D}_n \gets \mathcal{D}_{n-1}\cup\{a_n,\tilde J(a_n),\tilde G_1(a_n),\ldots,\tilde G_m(a_n)\}$
\EndFor
\For{$n = T_0+1,T_0+2,\ldots$}
    \State Fit/update the GP for $J$ and the GPs for all $G_i$ using $\mathcal{D}_{n-1}$
    \State Compute $\mathcal{S}_n \gets \{a\in\mathcal{A}\mid l_n^{(i)}(a)\ge h_i,\; i=1,\ldots,m\}$
    \State Select $a_n \gets \arg\max_{a\in\mathcal{S}_n} u_n^{(0)}(a)$
    \State Evaluate the controller with $a_n$ and observe $\tilde J(a_n),\tilde G_1(a_n),\ldots,\tilde G_m(a_n)$
    \State $\mathcal{D}_n \gets \mathcal{D}_{n-1}\cup\{a_n,\tilde J(a_n),\tilde G_1(a_n),\ldots,\tilde G_m(a_n)\}$
\EndFor
\end{algorithmic}
\end{algorithm}

\section{Theoretical Results}
\label{sec:theoretical_results}

This section states the guarantees used to justify the two stages of \textsc{SafeCtrlBO}. The guarantees follow the standard safe BO pattern: they hold on the GP confidence event in \Eqref{eq:confidence interval}, and therefore depend on valid choices of $\beta_n$, kernel hyperparameters, RKHS-norm bounds, and noise bounds. The boundary-based expansion rule also requires the geometric condition stated in Appendix~\ref{proof_thm3}; without that condition it should be interpreted as a computationally efficient heuristic, which we evaluate empirically in Appendix~\ref{sec:ablation}.

\begin{definition}[$\epsilon$-accurate safe expansion]
At the end of the expansion stage, the safe set is called $\epsilon$-accurate on a reachable safe region $\mathcal{R}_\epsilon$ if
\begin{equation}
    \max_{a\in\mathcal{R}_\epsilon}\max_{i=1,\ldots,m} 2\beta_t\sigma_{t-1}^{(i)}(a)\le \epsilon.
\end{equation}
Equivalently, every point in $\mathcal{R}_\epsilon$ has a safety-confidence width no larger than $\epsilon$.
\end{definition}

\begin{definition}[$\zeta$-optimal safe recommendation]
Let $a^*\in\arg\max_{a\in\mathcal{R}_\epsilon}J(a)$ be the best point in the safely reachable region. A recommendation $a_{\mathrm{rec}}\in\mathcal{R}_\epsilon$ is $\zeta$-optimal if $J(a^*)-J(a_{\mathrm{rec}})\le\zeta$.
\end{definition}

\begin{theorem}[Sufficient expansion budget]
\label{thm4}
Suppose the confidence event in \Eqref{eq:confidence interval} holds for all safety functions, each safety function $G_i$ is modeled with an additive kernel satisfying the RKHS and Lipschitz conditions in Appendix~\ref{sec:proof_additive_kernels}, and the boundary-expansion rule samples an $r$-cover of the reachable safe region $\mathcal{R}_\epsilon$. Let $\rho_\eta^2$ be an upper bound on the posterior variance that remains at an evaluated point because of observation noise; for noise-free or sufficiently averaged evaluations, $\rho_\eta=0$. Define
\begin{equation}
    \epsilon_\sigma = \sqrt{\left(\frac{\epsilon}{2\beta_{t^*}}\right)^2-\rho_\eta^2},
\end{equation}
and assume $\epsilon_\sigma>0$. For a first-order additive squared-exponential safety kernel with amplitude $\sigma_f$ and length-scales $\ell_j$, a sufficient coordinate-wise cover is
\begin{equation}
    r_j \le \frac{\ell_j\epsilon_\sigma}{\sigma_f\sqrt{D}}.
\end{equation}
Equivalently, it suffices to take
\begin{equation}
\label{eq:expansion-budget}
    t^* \ge \prod_{j=1}^{D}\left\lceil \frac{\sigma_f\sqrt{D}L_j}{\ell_j\epsilon_\sigma}\right\rceil,
\end{equation}
where $L_j$ is the side length of $\mathcal{R}_\epsilon$ in dimension $j$. Consequently, after $t^*$ expansion evaluations, every point in $\mathcal{R}_\epsilon$ has safety-confidence width at most $\epsilon$ with probability at least $1-\delta$. For higher-order additive kernels, the same statement holds after replacing the coordinate-wise bound by the corresponding kernel-metric cover induced by \Eqref{eq:kernel-metric-lipschitz}.
\end{theorem}

Theorem~\ref{thm4} shows a sufficient coverage condition. It makes explicit the dependency on the fill distance of the expansion samples and on the GP confidence width. The proof is given in Appendix~\ref{proof_thm4}.

\begin{theorem}[Simple-regret bound in the maximization stage]
\label{thm5}
Assume the expansion stage has identified a fixed safely reachable region $\mathcal{R}_\epsilon$ and the maximization stage evaluates only points in this region using \Eqref{eq:max-acquisition}. Suppose the objective $J$ has RKHS norm at most $B$, the observation noise is conditionally $R$-sub-Gaussian, and the confidence width is
\begin{equation}
    \beta_t = B + R\sqrt{2\left(\gamma_{t-1}+1+\ln\frac{1}{\delta}\right)},
\end{equation}
where $\gamma_t$ is the maximum information gain of the additive kernel. Let $a_{\mathrm{rec}}$ be the best objective value observed in the maximization stage after $T$ evaluations. On the confidence event,
\begin{equation}
\label{eq:simple-regret-bound}
    J(a^*)-J(a_{\mathrm{rec}}) \le 2\beta_T\sqrt{\frac{2\gamma_T}{T}}.
\end{equation}
Therefore, if $T=T^*$ satisfies
\begin{equation}
\label{eq:zeta-condition-general}
    \frac{8\beta_{T^*}^2\gamma_{T^*}}{T^*}\le \zeta^2,
\end{equation}
then the returned controller is $\zeta$-optimal with probability at least $1-\delta$.
\end{theorem}

For an additive squared-exponential kernel whose components have maximum interaction order $q_{\max}$, one may use an upper bound of the form $\gamma_T\le \bar\gamma(T)$, where $\bar\gamma(T)$ is the sum of the component-wise information-gain bounds. For example, if $M$ order-$q_{\max}$ components are used, a conservative squared-exponential bound is $\bar\gamma(T)=\mathcal{O}\!\left(M(\log T)^{q_{\max}+1}\right)$. Substituting such a bound into \Eqref{eq:zeta-condition-general} gives an explicit sufficient value of $T^*$. The proof of Theorem~\ref{thm5} is given in Appendix~\ref{proof_thm5}.


\section{Empirical Study}

\subsection{Simulations on Synthetic Benchmark Functions}
\label{sec:benchmark}

We first evaluate \textsc{SafeCtrlBO} on synthetic safe-optimization benchmarks where repeated trials are possible and the true optimum is known. The benchmarks are Camelback (2D), Hartmann (6D), and a Gaussian function (10D). We compare against representative safe BO baselines: \textsc{SwarmSafeOpt}, \textsc{SwarmStageOpt}, and the three \textsc{LineBO} variants from \citet{Johannes19}. We also include the high-dimensional additive BO method \textsc{DuMBO} \citep{dumbo} as an unconstrained reference to show the performance one can obtain when safety is ignored.

All benchmarks are converted to maximization problems by negating the standard minimization objective. We then impose synthetic safety thresholds. The inverted Camelback function has maximum value approximately $1.0316$, and the safety threshold is set to $0$. The inverted Hartmann function has maximum value approximately $3.32237$, and the safety threshold is set to $0.3$. For the 10D Gaussian benchmark, $f(x)=-\exp(-4\|x\|_2^2)$ before inversion, so the inverted maximum is $1$, and the safety threshold is set to $0.1$. A query whose value lies below the threshold is counted as a safety violation.

Each method is run for 100 random trials per benchmark. Every run starts from a randomly generated safe initial point. We use 150 iterations for Camelback and 200 iterations for Hartmann and Gaussian, which is intentionally smaller than the 400--1,000 iterations often used in high-dimensional line-search BO studies. This budget reflects hardware-control settings where long exploration is impractical. For \textsc{LineBO} and \textsc{DuMBO}, we use the public implementations and default hyperparameters when available. For \textsc{SwarmSafeOpt}, \textsc{SwarmStageOpt}, and \textsc{SafeCtrlBO}, we select hyperparameters manually and keep them fixed across the 100 runs. Additional implementation details are provided in Appendix~\ref{sec:appendix_implementation}.

\begin{figure*}[t]
    \centering
    
    \includegraphics[width=0.93\textwidth]{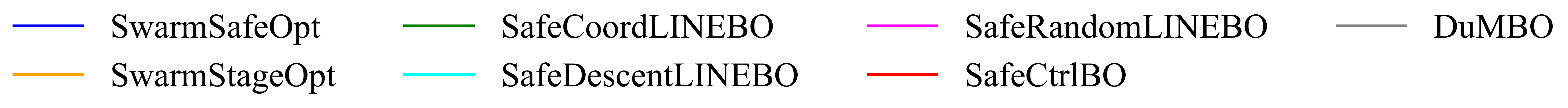}  
    \vspace{0.05cm}  
    
    \begin{tikzpicture}
    \setlength{\belowcaptionskip}{0pt}
        \node[anchor=south west,inner sep=0] (image) at (0,0) {
            \begin{tabular}{ccc}
                \hspace{-0.3cm}
                \begin{subfigure}[b]{0.32\textwidth}  
                    \includegraphics[width=\textwidth]{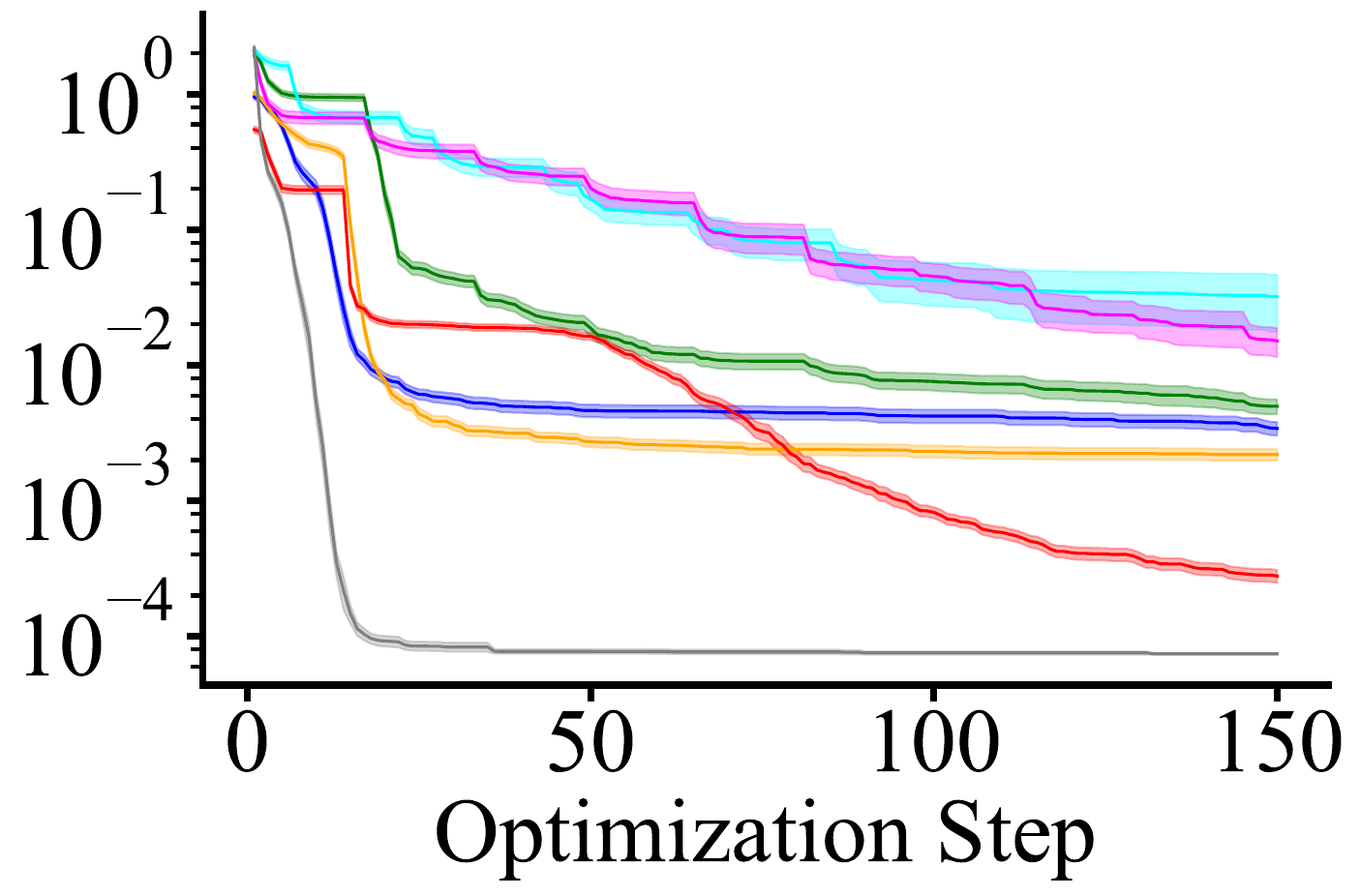}
                    \caption{Camelback2D}
                    \label{fig:camelback}
                \end{subfigure} 
                \hspace{-0.21cm} 
                \begin{subfigure}[b]{0.32\textwidth}  
                \includegraphics[width=\textwidth]{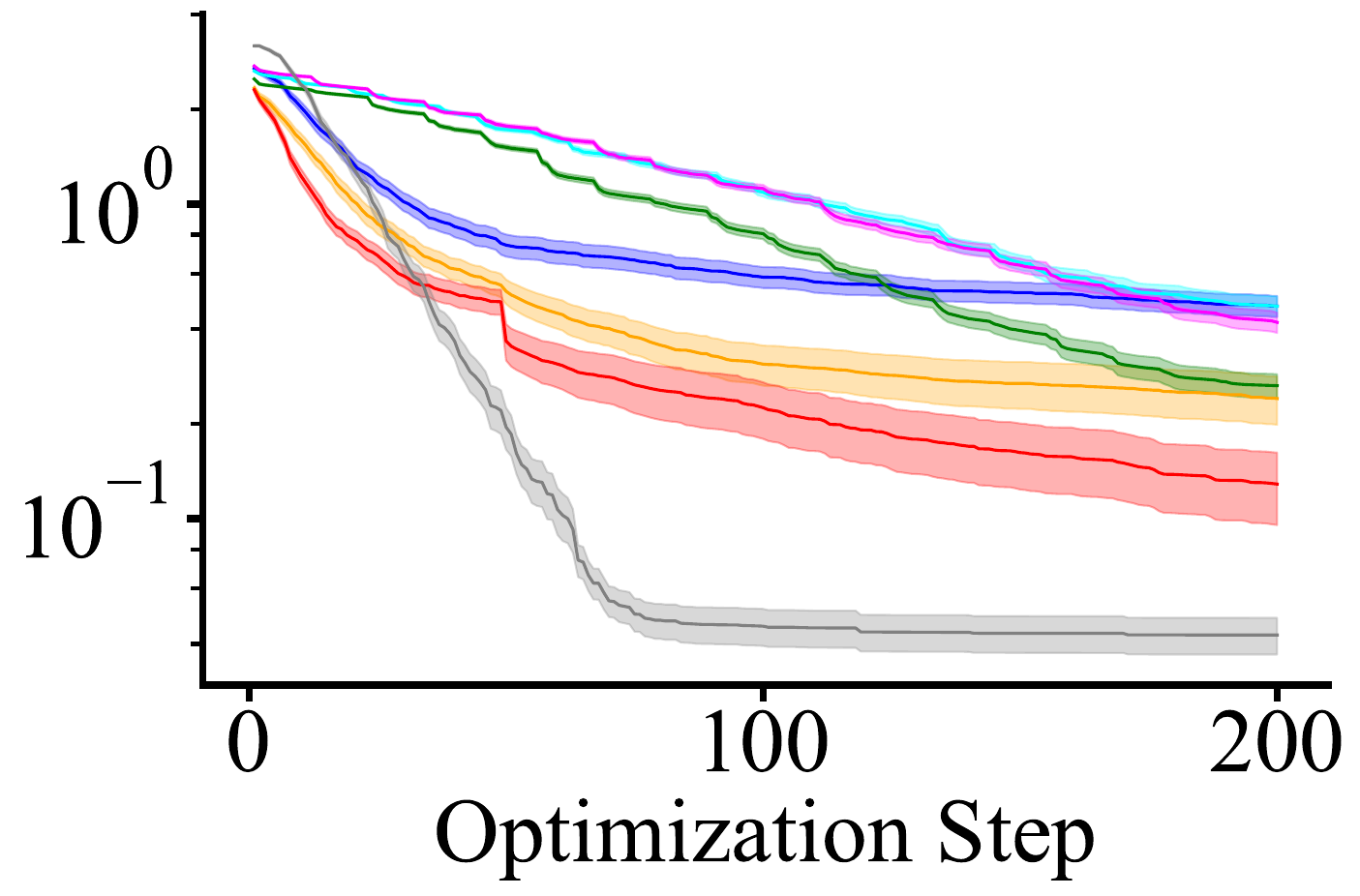}
                    \caption{Hartmann6D}
                    \label{fig:hartmann}
                \end{subfigure} 
                \hspace{-0.21cm}
                \begin{subfigure}[b]{0.32\textwidth}  

                    \includegraphics[width=\textwidth]{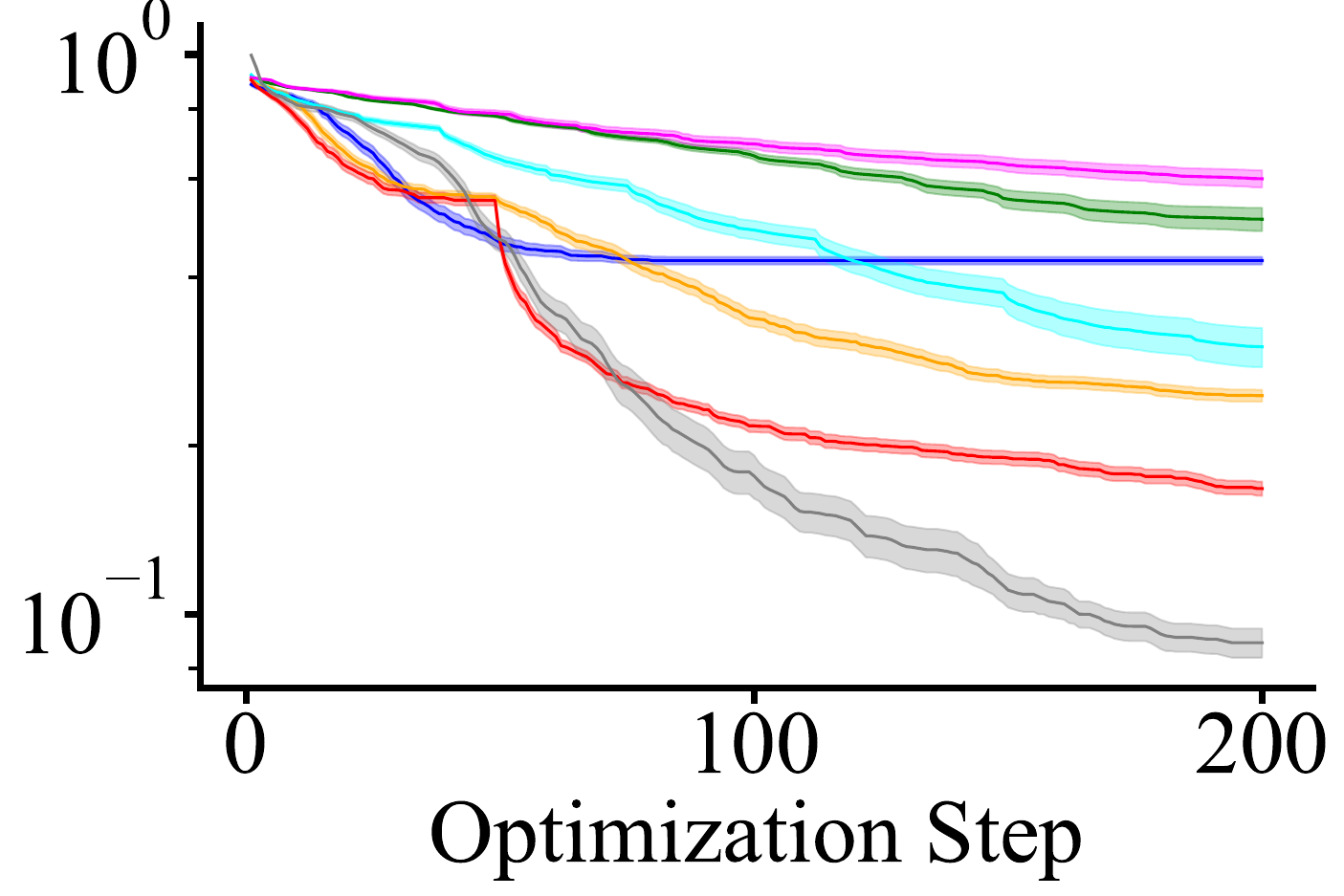}
                    \caption{Gaussian10D}
                    \label{fig:gaussian}
                \end{subfigure} \\
            \end{tabular}
        };
        \node[rotate=90, anchor=south] at ([yshift=0.7cm,font=\tiny]image.west) {\small{Simple Regret}};
    \end{tikzpicture}
    \caption{Optimization for synthetic benchmark functions.}
    \label{fig:synthetic simulations}
\end{figure*}

\paragraph{Results.} Figure~\ref{fig:synthetic simulations} reports the mean simple regret and standard error over 100 runs. As expected, unconstrained \textsc{DuMBO} achieves the lowest simple regret, but it violates the safety thresholds 1009, 3820, and 12,557 times on Camelback, Hartmann, and Gaussian, respectively. In contrast, the safe BO methods do not violate the synthetic safety thresholds under these experimental settings. With the limited iteration budgets, the \textsc{LineBO} variants improve slowly because they optimize along one-dimensional subspaces sequentially. \textsc{SwarmStageOpt} improves over \textsc{SwarmSafeOpt} by separating expansion and maximization, while \textsc{SafeCtrlBO} remains competitive and benefits from the additive surrogate in the 6D and 10D settings.

The regret curves also reveal the different optimization mechanisms. \textsc{LineBO} exhibits several segmented decreases as it moves from one subspace to another. \textsc{SwarmStageOpt} and \textsc{SafeCtrlBO} show a two-stage pattern caused by the switch from safe expansion to objective maximization. In these simulations, the switch occurs after 15 iterations for Camelback and after 50 iterations for Hartmann and Gaussian.

\subsection{Hardware Experiments on a Speedgoat Real-Time Machine}
\label{sec:experiment}

\begin{wrapfigure}{r}{7cm}
    \centering
    \includegraphics[width=7cm]{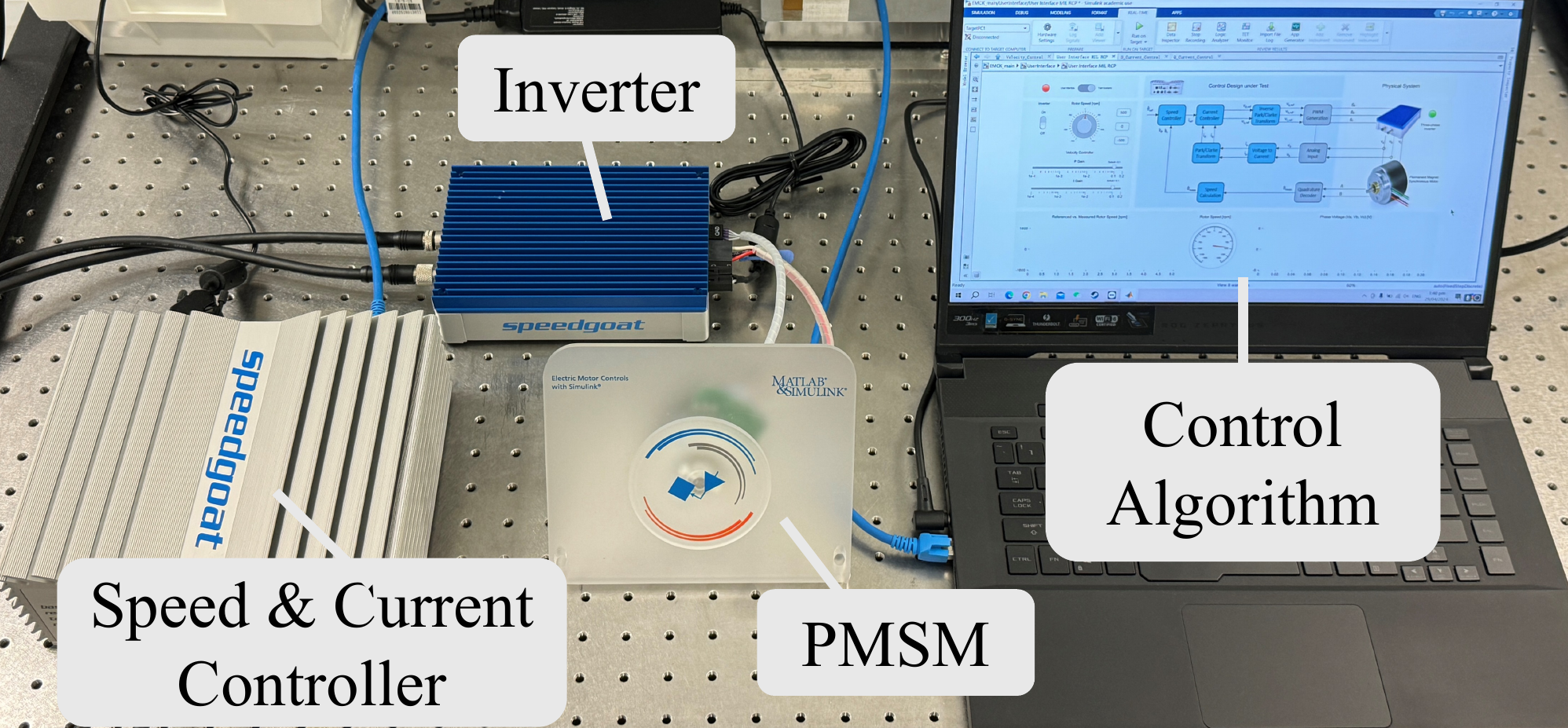}
    \caption{Hardware experimental setup.}
    \label{fig:setup}
\end{wrapfigure}
We next evaluate the methods on a physical PMSM platform. Unlike synthetic functions, repeated hardware trials with the same controller gains can produce slightly different measurements because of sensor noise, temperature, load variation, timing effects, and other unmodeled disturbances. This makes the experiment a useful test of robustness to real evaluation noise. The setup in Figure~\ref{fig:setup} contains a Speedgoat real-time controller with integrated speed and current loops, a Speedgoat inverter, and a PMSM. Appendix~\ref{sec:appendix_FOC_diagram} shows the corresponding FOC block diagram.

The FOC controller contains one outer speed PI controller and two inner current PI controllers for the $d$- and $q$-axes. The six gains are coupled: the speed-loop gains dominate speed tracking, the $q$-axis current loop strongly affects torque production, and the $d$-axis current loop affects current and flux behavior. Tuning the loops independently can therefore miss high-performing combinations, while unconstrained joint search can produce excessive current or unacceptable tracking behavior.

The objective is to improve speed tracking by reducing settling time, overshoot, and steady-state error. We use the performance score
\begin{equation}
    J(t_s,O_s,e_{ss}) = w_s(t_0-t_s)-w_o O_s-w_e e_{ss},
\end{equation}
where $t_s$ is the $2\%$ settling time, $O_s$ is overshoot, and $e_{ss}$ is steady-state error. The weights are $w_s=20$, $w_o=1.5$, $w_e=4$, and $t_0=2.5$.

\begin{figure*}[t]
    \centering
    \includegraphics[width=0.99\textwidth]{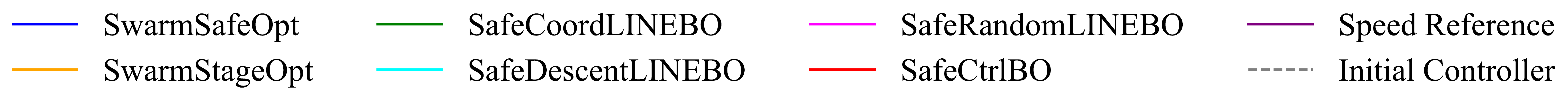}
    \vspace{0.08cm}
    \begin{tikzpicture}
    \setlength{\belowcaptionskip}{0pt}
        \node[anchor=south west,inner sep=0] (image) at (0,0) {
            \begin{tabular}{ccc}
                \hspace{-0.3cm}
                \begin{subfigure}[b]{0.43\textwidth}
                    \includegraphics[width=\textwidth]{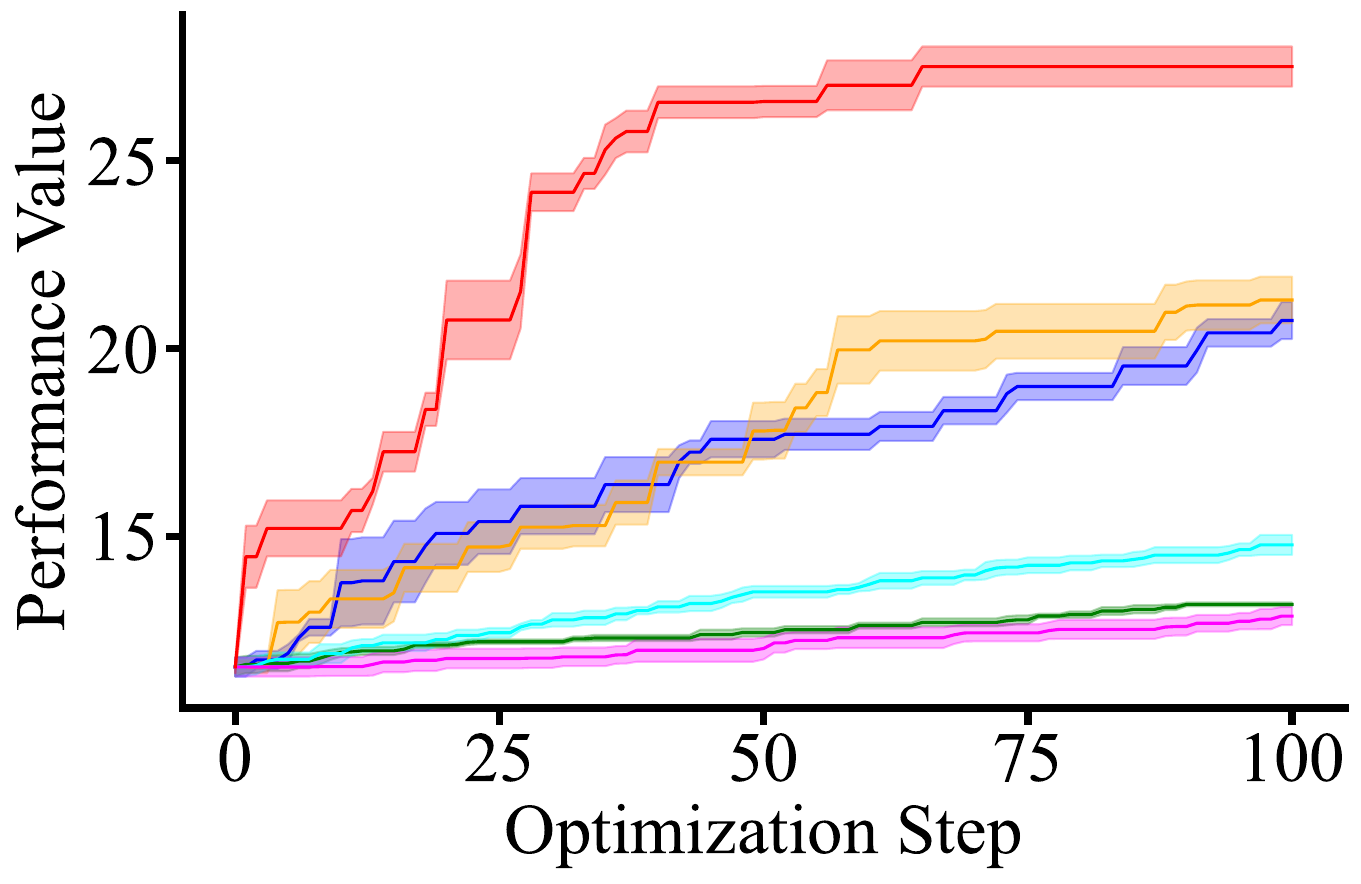}
                    \caption{Comparison of performance values.}
                    \label{fig:experiment1}
                \end{subfigure}
                \hspace{0.25cm}
                \begin{subfigure}[b]{0.44\textwidth}
                    \includegraphics[width=\textwidth]{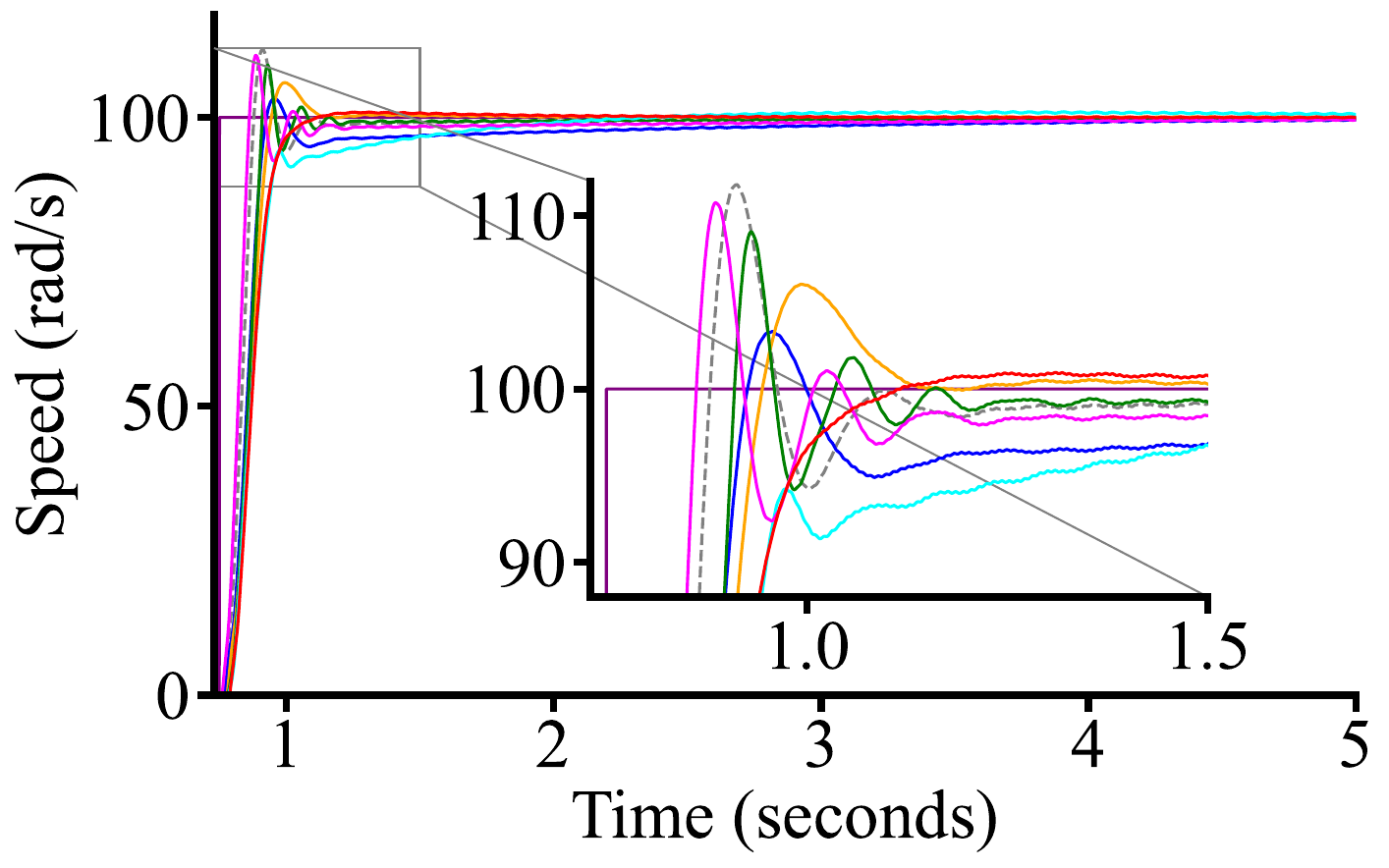}
                    \caption{Comparison of speed tracking curves.}
                    \label{fig:experiment3}
                \end{subfigure} \\
            \end{tabular}
        };
    \end{tikzpicture}
    \caption{Hardware experiment results.}
    \label{fig:hardware experiment}
\end{figure*}

We use two explicit safety functions in addition to the performance threshold: one discourages large steady-state error and the other limits the control signal energy,
\begin{equation}
\begin{aligned}
    G_e &= C_{e0}-w_e' e_{ss}, \\
    G_u &= C_{u0} - w_u \sum_{t=0}^{1} u(t)^2.
\end{aligned}
\end{equation}
The constants are $C_{e0}=C_{u0}=100$, $w_e'=40$, and $w_u=0.001$, and both thresholds are set to $0$. The $G_u$ constraint is the hard signal-safety constraint; the performance and steady-state-error thresholds are softer engineering constraints used to guide search away from poor controllers. The initial controller supplied by the Speedgoat model satisfies these thresholds.

The six controller gains have different ranges: the speed-controller $p$ and $i$ gains lie in $[0.01,0.5]$; the $d$- and $q$-axis current-controller $p$ gains lie in $[0.1,1]$; and the current-controller $i$ gains lie in $[1,200]$. These different scales motivate dimension-wise lengthscales and variances in the additive kernel. They also motivate including interaction terms, because the speed loop and current loops influence different but coupled parts of the tracking response.

For safety reasons, we compare only safe BO algorithms on hardware: \textsc{SwarmSafeOpt}, \textsc{SwarmStageOpt}, the three \textsc{LineBO} variants, and \textsc{SafeCtrlBO}. The \textsc{LineBO} implementation used here only supports the performance function as its safety metric, so $G_e$ and $G_u$ are not evaluated for \textsc{LineBO}. Each method is run five times for 100 iterations. Figure~\ref{fig:experiment1} reports the mean and standard error of the performance values, Figure~\ref{fig:experiment3} shows the best speed-tracking curve found by each method, and Table~\ref{tab:comparison} summarizes the corresponding metrics.

\paragraph{Results.} Figure~\ref{fig:experiment1} shows that all methods improve performance over the initial controller, but \textsc{SafeCtrlBO} improves fastest and reaches the highest best value. Table~\ref{tab:comparison} shows that the best controller found by \textsc{SafeCtrlBO} has the smallest overshoot, the shortest $2\%$ settling time, and the second-smallest steady-state error. For constraint violations, we distinguish soft threshold violations from hard signal-safety violations. The three \textsc{LineBO} variants only monitor the performance threshold; \textsc{SafeCoordLineBO}, \textsc{SafeDescentLineBO}, and \textsc{SafeRandomLineBO} violate that threshold 1, 4, and 5 times, respectively. \textsc{SwarmSafeOpt}, \textsc{SwarmStageOpt}, and \textsc{SafeCtrlBO} monitor the performance threshold, $G_e$, and $G_u$; across five runs they record 39, 61, and 39 total threshold violations, respectively. As discussed in Appendix~\ref{sec:hardware_constraint_violation}, these violations are dominated by soft performance or steady-state-error thresholds rather than hard signal-safety failures.

In this experiment, \textsc{SafeCtrlBO} uses a full six-order additive kernel. This expressive kernel improves optimization but increases the cost of evaluating the exact potential-expander set. With the potential-expander calculation used in previous safe BO methods, one iteration takes approximately 48 seconds. With the boundary set $\mathcal{B}_n(\tau_n)$, the iteration time decreases to approximately 28 seconds. The boundary simplification is therefore important not only for sample efficiency, but also for wall-clock efficiency during hardware tuning.

\begin{table}[t]
    \caption{Performance comparison of the best PMSM speed tracking curves optimized by different methods. The best results are written in \textbf{bold} text, and the second-best results are \underline{underlined}.}
    \label{tab:comparison}
    \centering
    \begin{tabular}{c c c c c}
        \toprule
         Method         & $J(a^*)$ $\uparrow$ & $O_s$ $(rad/s)$ $\downarrow$ & $e_{ss}$ $(rad/s)$ $\downarrow$ & $2\% \: t_s$ $(s)$ $\downarrow$    \\
        \midrule
        Initial Controller    & 11.6788 & 11.789 & 0.067 & \underline{0.301}         \\
        \textsc{SwarmSafeOpt} & 21.4277 & 3.323 & 0.417 & 1.721        \\
        \textsc{SwarmStageOpt} & \underline{23.7091} & 6.063 & \textbf{0.030} & 0.327       \\
        \textsc{SafeCoordLineBO} & 13.2615 & 9.743 & 0.081 & 0.431   \\
        \textsc{SafeDescentLineBO} & 14.734 & \underline{1.028} & 1.681 & 0.988     \\
        \textsc{SafeRandomLineBO} & 12.2435 & 10.747 & 0.409 & 0.477      \\
        \textsc{SafeCtrlBO} & \textbf{28.7893} & \textbf{0.956} & \underline{0.037} & \textbf{0.284}            \\
        \bottomrule
    \end{tabular}
\end{table}




\section{Conclusion}
\label{sec:conclusion}

This paper presented \textsc{SafeCtrlBO}, a safe Bayesian optimization method for tuning multiple coupled controllers under limited hardware-evaluation budgets. The method combines additive GP surrogates with a stagewise safe BO procedure. The additive kernel gives the surrogate an inductive bias toward main effects and low-order interactions among controller gains, while the boundary-based expansion rule reduces the computational cost of safe-set expansion. Synthetic benchmarks and PMSM hardware experiments show that the method is competitive with representative safe BO baselines and is particularly useful when the number of permissible evaluations is small.

The main limitation is scalability of the additive kernel design. Including all interaction orders becomes expensive as the parameter dimension grows, and selecting useful interaction orders still requires either domain knowledge or an additional model-selection step. Another limitation is shared by most GP-based safe BO methods: the theoretical safety statement depends on calibrated confidence intervals, valid noise assumptions, and a meaningful RKHS-norm or regularity bound. In future work, kernel selection methods \citep{kernel_align_1,kernel_align_2,kernel_align_3,zheng2024robotic}, sparse or approximate GP inference, and safety filters based on control-theoretic constraints could be combined with \textsc{SafeCtrlBO} to improve scalability and practical safety.


\subsubsection*{Acknowledgments}
This work was supported by RIE2025 Manufacturing, Trade and Connectivity (MTC) Industry Alignment Fund – Pre-Positioning (IAF-PP) under Grant M22K4a0044 through WP3-Energy Efficient Motor Drive System with GaN-based Traction Inverters.

\bibliography{iclr2025_conference}
\bibliographystyle{iclr2025_conference}

\newpage
\appendix

\section{Architecture of the Field-Oriented Control Scheme on a PMSM}
\label{sec:appendix_FOC_diagram}

\begin{figure}[H]
    \centering
    \includegraphics[width=\linewidth]{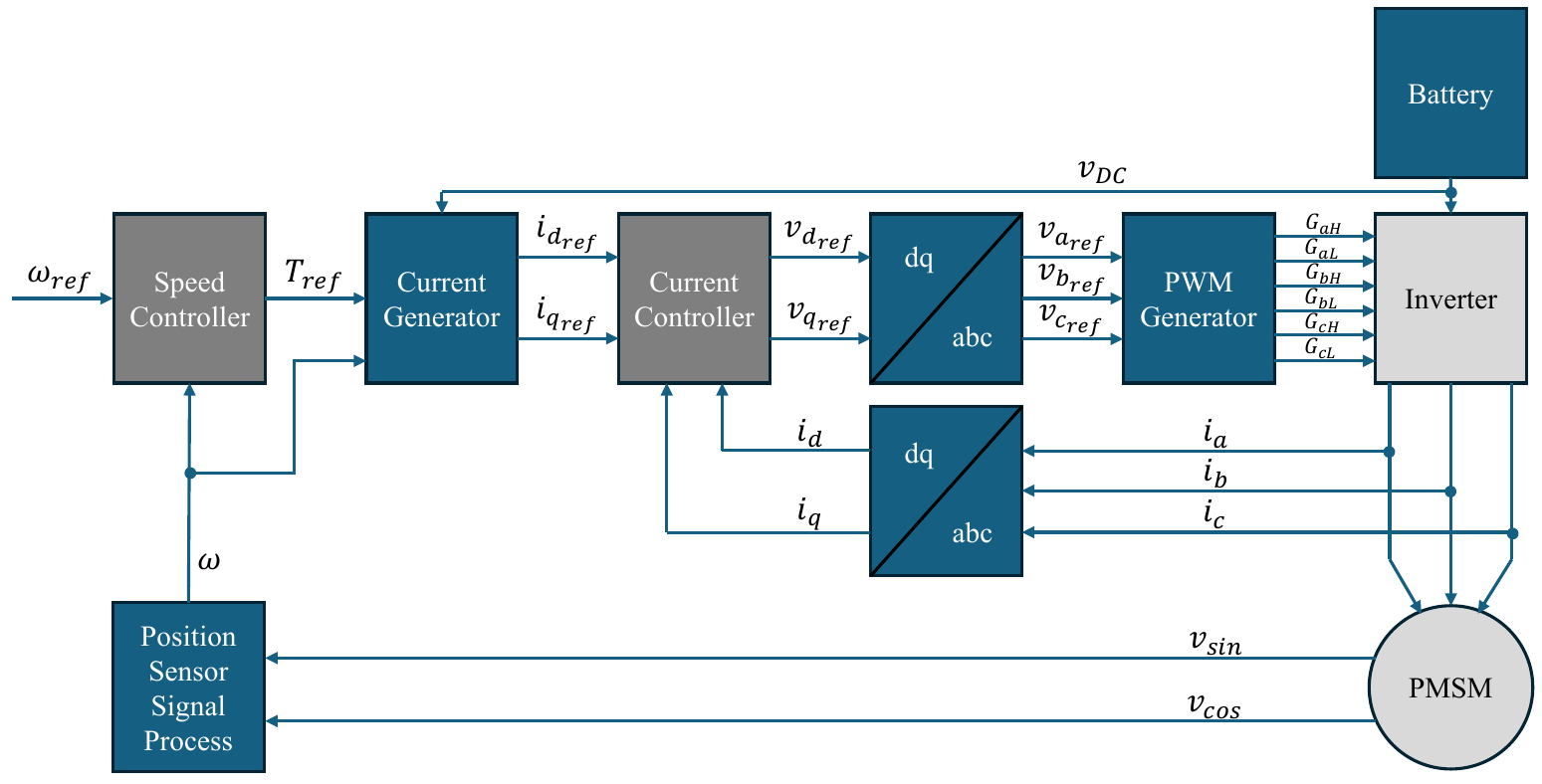}
    \caption{A simplified block diagram for PMSM FOC loops. The dark grey blocks represent controllers, and the light grey blocks represent plants.}
    \label{fig:PMSM}
\end{figure}

\section{Implementation Details of Experiments}
\label{sec:appendix_implementation}

The public implementation of \textsc{SafeCtrlBO} is available at \url{https://github.com/hxwangnus/SafeCtrlBO}. The experiments reported in this manuscript were conducted with an earlier GPy-based implementation, while the released repository has been refactored to GPyTorch. This dependency change does not alter the algorithmic description in Algorithm~\ref{algorithm}; it only changes the GP software backend used to fit the surrogate models.

\section{Discussion on Constraint Violation in Hardware Experiments}
\label{sec:hardware_constraint_violation}

\subsection{Discussion on Safety Guarantee}
In the hardware experiments, \textsc{SafeCtrlBO} records 39 threshold violations over five runs. These should not all be interpreted as hardware-safety failures. Most violations come from the performance threshold, which is a soft acceptability constraint: a violation means that the trial response is slower or has larger overshoot than desired. Setting this threshold to $-\infty$ would remove these violations, but it would also remove useful guidance that steers the optimizer away from poor controllers.

A smaller number of violations are associated with the steady-state-error constraint $G_e$. These indicate that the motor speed settles farther from the desired setpoint than preferred. This is also an engineering-performance constraint rather than a direct hardware-damage constraint. It is included because controllers with large steady-state error are usually not useful, and excluding them improves the efficiency of the search.

The hard safety constraint in this experiment is the signal-safety constraint $G_u$, which limits the control signal energy and thereby protects the hardware from excessive current. The reported experiments did not observe damaging behavior under this constraint. Nevertheless, the guarantee provided by GP-based safe BO is a high-probability guarantee, not an absolute guarantee. It depends on calibrated confidence intervals, valid noise assumptions, and a conservative enough initial safe set.

\subsection{Discussion on the Safety--Efficiency Trade-off}
The threshold violations occur primarily during the expansion stage, when the algorithm intentionally probes the boundary of the current safe set to learn where the safe region can grow. A more conservative choice of confidence width, kernel lengthscale, or safety threshold can reduce violations, but it will typically require more evaluations. This trade-off is unavoidable in hardware tuning: stronger practical safety margins slow down exploration, while aggressive exploration improves sample efficiency but increases the chance of soft-threshold violations.

As noted in the safe BO literature, confidence-interval methods certify safety with high probability rather than with probability one \citep{Sui18}. Recent analyses further emphasize that practical safety depends on whether the implemented uncertainty bounds are valid for the true system and noise model \citep{fiedler2024safety}. For safety-critical platforms, \textsc{SafeCtrlBO} should therefore be combined with conservative physical interlocks, emergency shutdown conditions, or model-based safety filters.

\subsection{Practical Guidance}
The appropriate balance between safety and efficiency depends on the platform. If each trial is cheap and wear is negligible, conservative thresholds and larger $\beta_n$ values are preferable. If each trial is expensive but the hardware has robust low-level protection, softer performance thresholds can accelerate search while the hard physical constraints remain conservative. For platforms such as drones, where an unsafe controller can crash the system, the safe BO layer should not be the only protection mechanism. For motor-drive experiments such as the PMSM study here, poor tracking responses can often be stopped by software or human supervision, while current and voltage limits should remain hard constraints.

\section{Detailed Proofs}
\label{sec:proofs}

\subsection{Proofs for the Properties of Additive Gaussian Kernels}
\label{sec:proof_additive_kernels}

\renewcommand\qedsymbol{$\blacksquare$}

\begin{lemma}
\label{lemma1}
Let $\{k_I\}_{I\in\mathcal{I}}$ be positive definite component kernels on subspaces indexed by $I\subseteq\{1,\ldots,D\}$, and let $\lambda_I\ge 0$. Then
\begin{equation}
    k_{\mathrm{add}}(a,a')=\sum_{I\in\mathcal{I}}\lambda_I k_I(a_I,a'_I)
\end{equation}
is positive definite. If $\lambda_I>0$, the corresponding RKHS consists of functions that can be written as $f(a)=\sum_{I\in\mathcal{I}} f_I(a_I)$ with $f_I\in\mathcal{H}_{k_I}$, completed under the norm
\begin{equation}
\label{eq:additive-rkhs-norm}
    \|f\|_{k_{\mathrm{add}}}^2
    = \inf_{f=\sum_I f_I}\sum_{I\in\mathcal{I}}\frac{\|f_I\|_{k_I}^2}{\lambda_I}.
\end{equation}
\end{lemma}

\begin{proof}
Positive definiteness follows from closure of positive definite kernels under nonnegative weighted sums. For any finite set $\{a_1,\ldots,a_n\}$ and vector $\alpha\in\mathbb{R}^n$,
\begin{equation}
    \sum_{r,s=1}^n \alpha_r\alpha_s k_{\mathrm{add}}(a_r,a_s)
    = \sum_{I\in\mathcal{I}} \lambda_I \sum_{r,s=1}^n \alpha_r\alpha_s k_I((a_r)_I,(a_s)_I) \ge 0.
\end{equation}
The RKHS characterization in \Eqref{eq:additive-rkhs-norm} is the standard RKHS sum construction: kernel sections decompose as $k_{\mathrm{add}}(\cdot,a)=\sum_I \lambda_I k_I(\cdot_I,a_I)$, and the inner product is the minimum-energy decomposition over component RKHSs. This gives the reproducing property
\begin{equation}
    f(a)=\langle f,k_{\mathrm{add}}(\cdot,a)\rangle_{k_{\mathrm{add}}}.
\end{equation}
\end{proof}

\begin{theorem}[Bounded RKHS norm under an additive decomposition]
\label{thm1}
Assume $f(a)=\sum_{I\in\mathcal{I}} f_I(a_I)$ with $f_I\in\mathcal{H}_{k_I}$ and $\|f_I\|_{k_I}\le B_I$. If all $\lambda_I>0$, then
\begin{equation}
    \|f\|_{k_{\mathrm{add}}}^2 \le \sum_{I\in\mathcal{I}}\frac{B_I^2}{\lambda_I}.
\end{equation}
\end{theorem}

\begin{proof}
The claimed decomposition is one feasible decomposition in the infimum defining the additive RKHS norm in \Eqref{eq:additive-rkhs-norm}. Therefore,
\begin{equation}
    \|f\|_{k_{\mathrm{add}}}^2
    \le \sum_{I\in\mathcal{I}}\frac{\|f_I\|_{k_I}^2}{\lambda_I}
    \le \sum_{I\in\mathcal{I}}\frac{B_I^2}{\lambda_I}.
\end{equation}
\end{proof}

Theorem~\ref{thm1} should be read as a modeling assumption: the target function must be well represented by the selected additive components. It does not imply that every arbitrary $D$-dimensional function has a small norm in a low-order additive RKHS.

\begin{theorem}[Lipschitz continuity of bounded-norm additive RKHS functions]
\label{thm2}
Suppose the additive kernel satisfies the kernel-metric bound
\begin{equation}
\label{eq:kernel-metric-lipschitz}
    \sqrt{k_{\mathrm{add}}(a,a)+k_{\mathrm{add}}(a',a')-2k_{\mathrm{add}}(a,a')} \le L_k\|a-a'\|_2
\end{equation}
for all $a,a'\in\mathcal{A}$. Then every $f\in\mathcal{H}_{k_{\mathrm{add}}}$ with $\|f\|_{k_{\mathrm{add}}}\le B$ is $BL_k$-Lipschitz:
\begin{equation}
    |f(a)-f(a')|\le BL_k\|a-a'\|_2.
\end{equation}
For squared-exponential additive components on a compact domain, such an $L_k<\infty$ exists; for first-order components, one may take $L_k^2\le \sum_{j=1}^D \lambda_j\sigma_{f,j}^2/\ell_j^2$.
\end{theorem}

\begin{proof}
By the reproducing property and Cauchy--Schwarz,
\begin{align}
    |f(a)-f(a')|
    &= |\langle f,k_{\mathrm{add}}(\cdot,a)-k_{\mathrm{add}}(\cdot,a')\rangle_{k_{\mathrm{add}}}| \\
    &\le \|f\|_{k_{\mathrm{add}}}\,\|k_{\mathrm{add}}(\cdot,a)-k_{\mathrm{add}}(\cdot,a')\|_{k_{\mathrm{add}}}.
\end{align}
The squared norm of the difference between kernel sections is
\begin{equation}
    \|k_{\mathrm{add}}(\cdot,a)-k_{\mathrm{add}}(\cdot,a')\|_{k_{\mathrm{add}}}^2
    = k_{\mathrm{add}}(a,a)+k_{\mathrm{add}}(a',a')-2k_{\mathrm{add}}(a,a').
\end{equation}
Using \Eqref{eq:kernel-metric-lipschitz} and $\|f\|_{k_{\mathrm{add}}}\le B$ gives the result. For one-dimensional squared-exponential components,
\begin{equation}
    2\lambda_j\sigma_{f,j}^2\left(1-\exp\left[-\frac{(a_j-a'_j)^2}{2\ell_j^2}\right]\right)
    \le \lambda_j\frac{\sigma_{f,j}^2}{\ell_j^2}(a_j-a'_j)^2,
\end{equation}
where we used $1-e^{-x}\le x$. Summing over dimensions yields the stated first-order bound. Higher-order components are smooth on compact domains, so an analogous finite bound follows by the same kernel-metric argument; see also \citet{fiedler2023lipschitz} for general RKHS regularity conditions.
\end{proof}

\subsection{Proofs for Simplifying the Safe Expansion Stage}
\label{proof_thm3}

The boundary-expansion rule used by \textsc{SafeCtrlBO} is a computational simplification of the potential-expander search in \textsc{SafeOpt}. The key point is not that the posterior variance of an RBF GP is globally monotone away from the data---this is false in general for multi-point posteriors---but that the simplification is valid under a local outward-variance condition that is often observed during safe-set expansion.

Let $C_n$ be the set of evaluated safe points at iteration $n$, let $O_n$ denote the portion of the candidate domain that can be connected to $C_n$ by outward segments, and let $\mathcal{B}_n$ be the corresponding safe-boundary set. For $x_{\mathrm{in}}\in C_n$ and $x_{\mathrm{b}}\in\mathcal{B}_n$, define the segment
\begin{equation}
    x(\lambda)=x_{\mathrm{in}}+\lambda(x_{\mathrm{b}}-x_{\mathrm{in}}),\qquad \lambda\in[0,1].
\end{equation}

\begin{assumption}[Outward variance monotonicity]
\label{assump:outward_variance}
For every segment used to parameterize $O_n$, the safety-posterior variance is nondecreasing along the outward direction:
\begin{equation}
    \frac{d}{d\lambda}\sigma_{n-1}^{(i)}(x(\lambda))\ge 0,
    \qquad \lambda\in[0,1],\quad i=1,\ldots,m .
\end{equation}
For a finite candidate set, the same condition is understood in the discrete sense: points farther along the segment have no smaller posterior standard deviation.
\end{assumption}

\begin{proposition}[Boundary maximizer under outward variance monotonicity]
\label{prop:boundary_max}
Under Assumption~\ref{assump:outward_variance}, for every safety function $G_i$,
\begin{equation}
    \max_{x\in O_n}\sigma_{n-1}^{(i)}(x)
    =\max_{x\in\mathcal{B}_n}\sigma_{n-1}^{(i)}(x).
\end{equation}
Consequently, selecting the most uncertain safe-boundary point gives the same expansion candidate as selecting the most uncertain point over $O_n$. If the relaxed boundary set $\mathcal{B}_n(\tau_n)$ in \Eqref{eq:boundary-set} is used, the selected point converges to a boundary maximizer as $\tau_n\to0$ and the candidate discretization is refined.
\end{proposition}

\begin{proof}
By construction, each $x\in O_n$ lies on at least one segment $x(\lambda)$ whose endpoint $x(1)$ belongs to $\mathcal{B}_n$. Assumption~\ref{assump:outward_variance} implies
\begin{equation}
    \sigma_{n-1}^{(i)}(x(\lambda))\le \sigma_{n-1}^{(i)}(x(1)),
    \qquad \lambda\in[0,1].
\end{equation}
Therefore the maximum posterior standard deviation on every segment is attained at its boundary endpoint. Taking the union over all such segments proves the equality of the maxima over $O_n$ and $\mathcal{B}_n$. The statement for $\mathcal{B}_n(\tau_n)$ follows because the relaxed set is an outer approximation of the level-set boundary and shrinks to $\mathcal{B}_n$ as $\tau_n$ and the grid resolution go to zero.
\end{proof}

\paragraph{When the condition is reasonable.}
For a single noiseless observation and an isotropic squared-exponential kernel, the posterior variance has the closed form
\begin{equation}
    \sigma^2(x)=\sigma_f^2\left(1-\exp\left[-\frac{\|x-x_0\|_2^2}{\ell^2}\right]\right),
\end{equation}
which increases with the distance from the observed point. With multiple observations, however, the derivative
\begin{equation}
    \frac{d}{d\lambda}\sigma_{n-1}^2(x(\lambda))
    =-2\left(\frac{d k_{n-1}(x(\lambda))}{d\lambda}\right)^{\!\top}
    (K_{n-1}+\sigma_\eta^2 I)^{-1}k_{n-1}(x(\lambda))
\end{equation}
need not be nonnegative for every direction. This is why the main text presents boundary search as a sufficient-condition result and a practical approximation rather than an unconditional theorem. The following visual checks show that the approximation recovers the same acquisition behavior as potential-expander search in the low-dimensional cases used to motivate the simplification.

\begin{figure}[ht]
     \centering
     \begin{subfigure}[b]{0.32\textwidth}
         \centering
         \includegraphics[width=\textwidth]{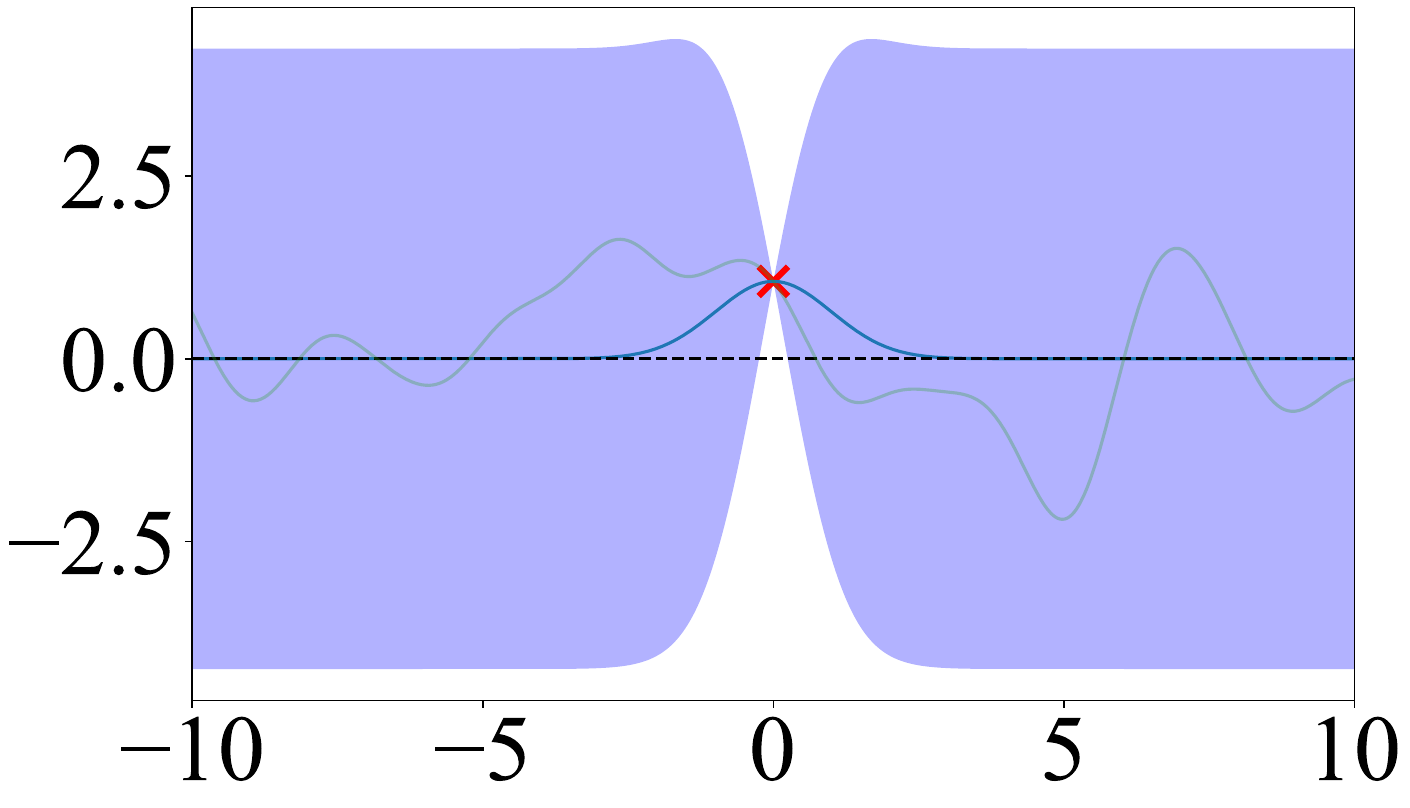}
         \caption{Initial safe prior.}
     \end{subfigure}
     \begin{subfigure}[b]{0.32\textwidth}
         \centering
         \includegraphics[width=\textwidth]{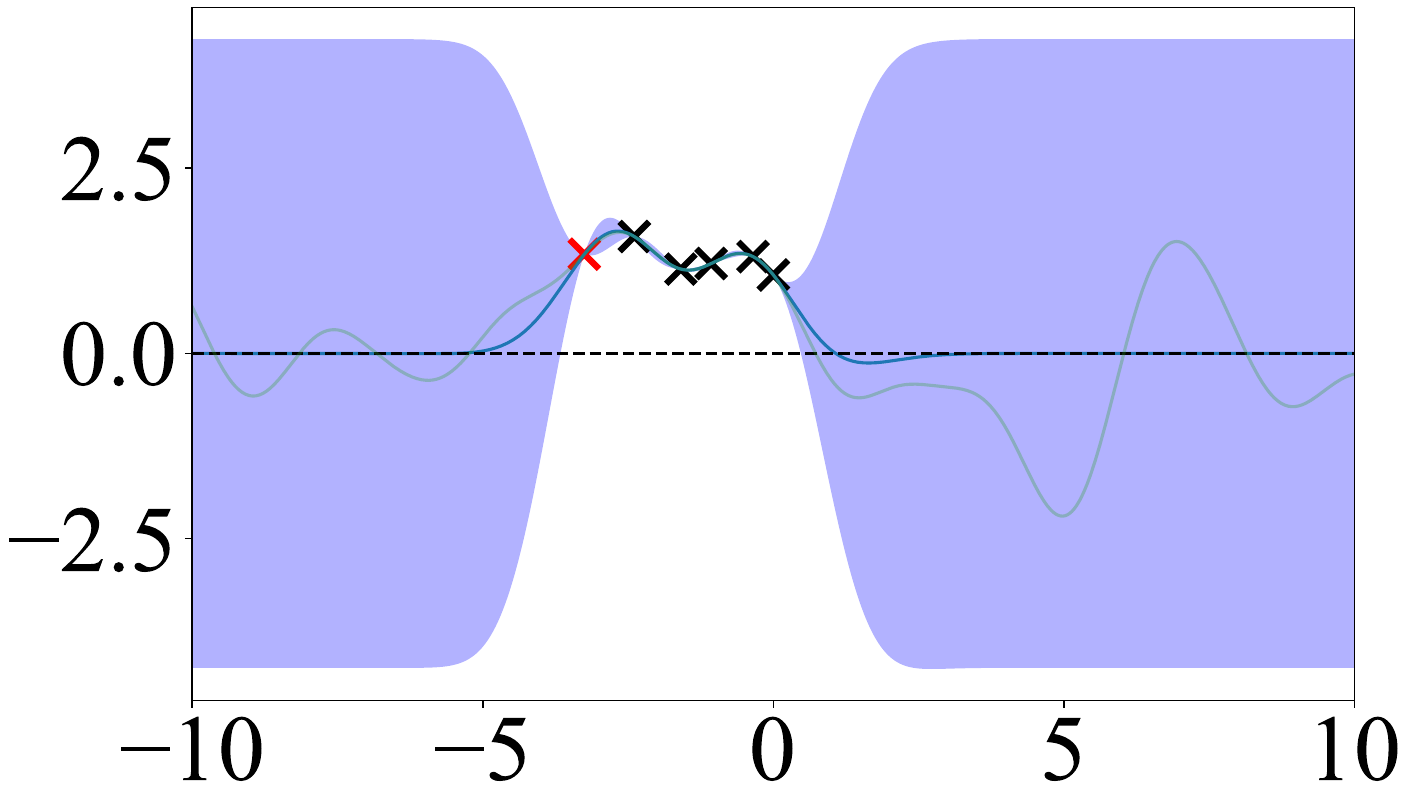}
         \caption{After 5 iterations.}
     \end{subfigure}
     \begin{subfigure}[b]{0.32\textwidth}
         \centering
         \includegraphics[width=\textwidth]{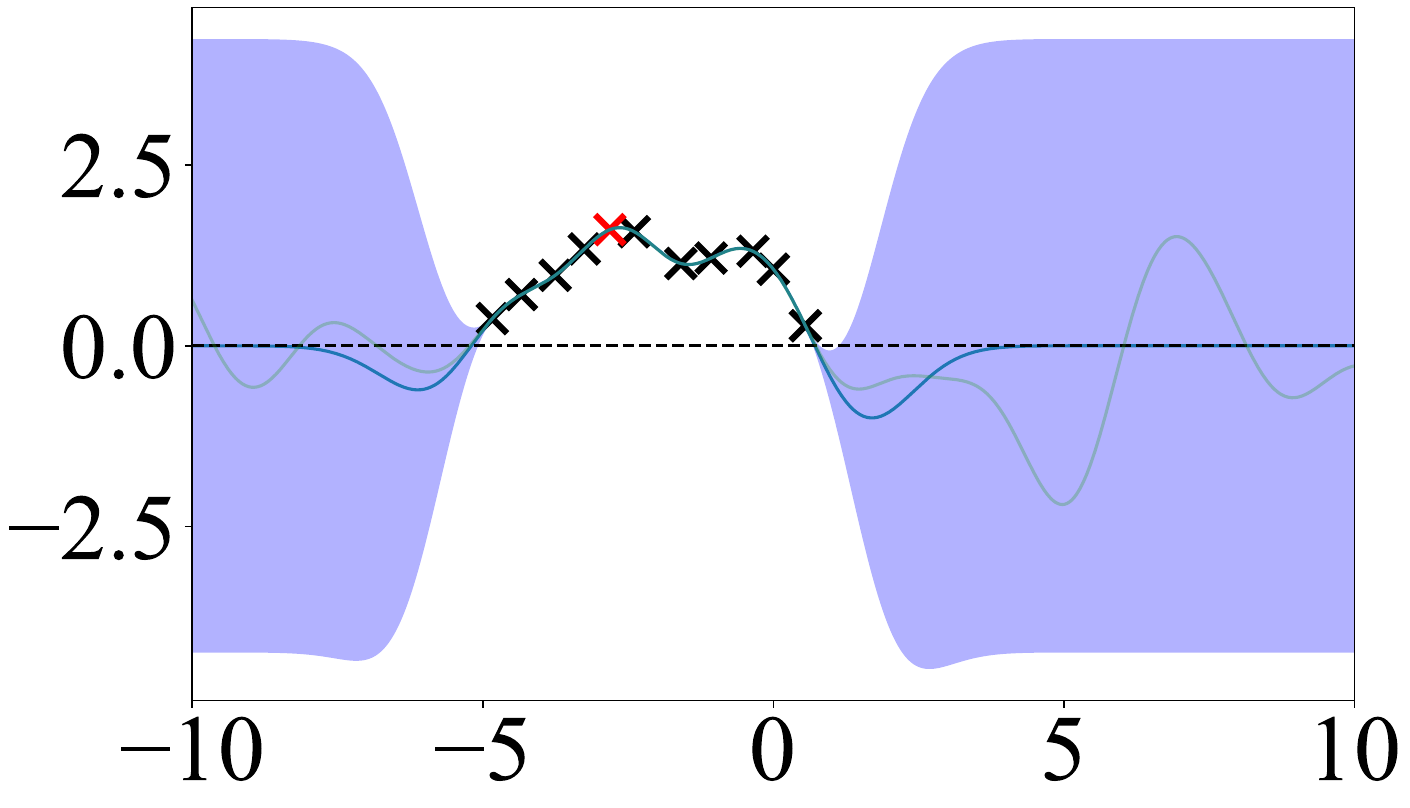}
         \caption{After 10 iterations.}
     \end{subfigure}
     \begin{subfigure}[b]{0.32\textwidth}
         \centering
         \includegraphics[width=\textwidth]{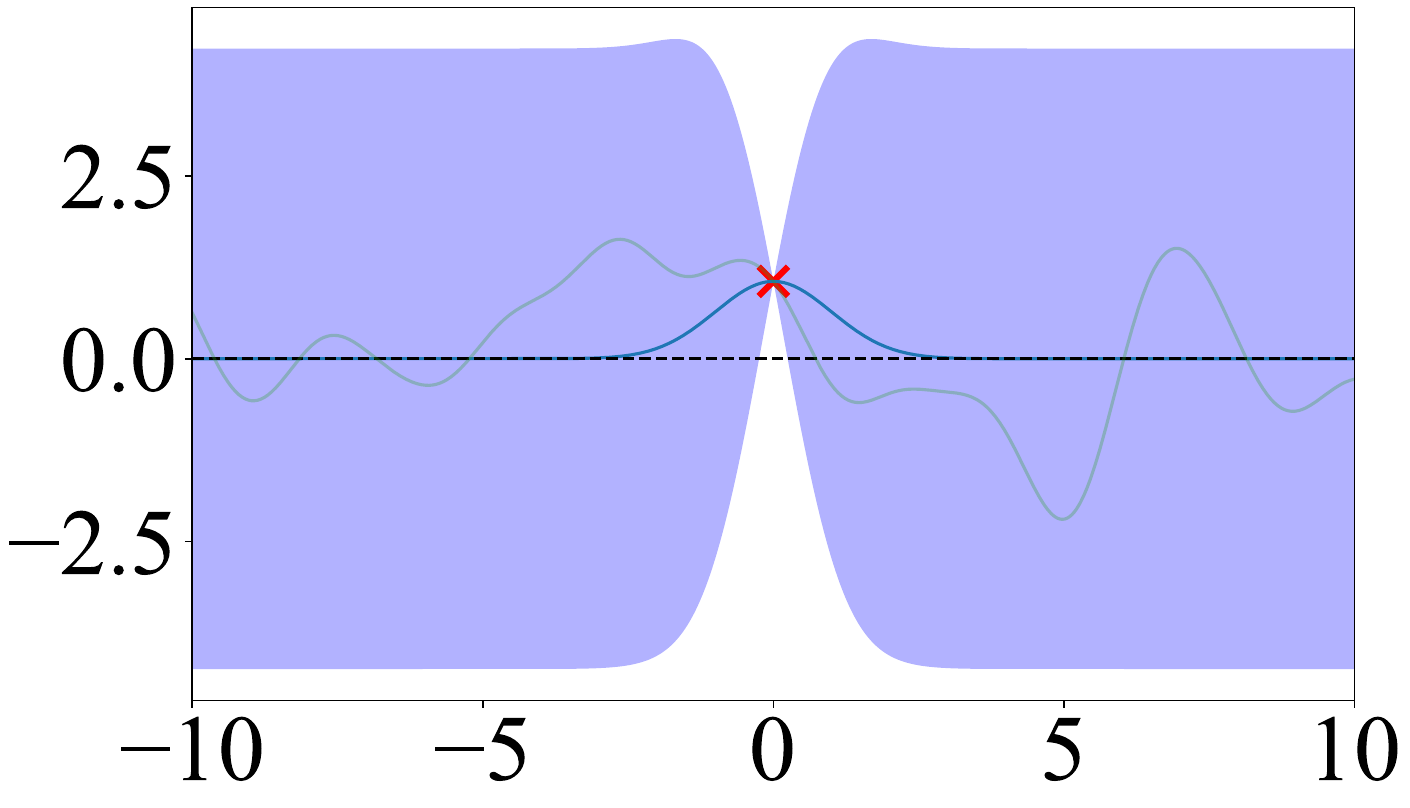}
         \caption{Initial safe prior.}
     \end{subfigure}
     \begin{subfigure}[b]{0.32\textwidth}
         \centering
         \includegraphics[width=\textwidth]{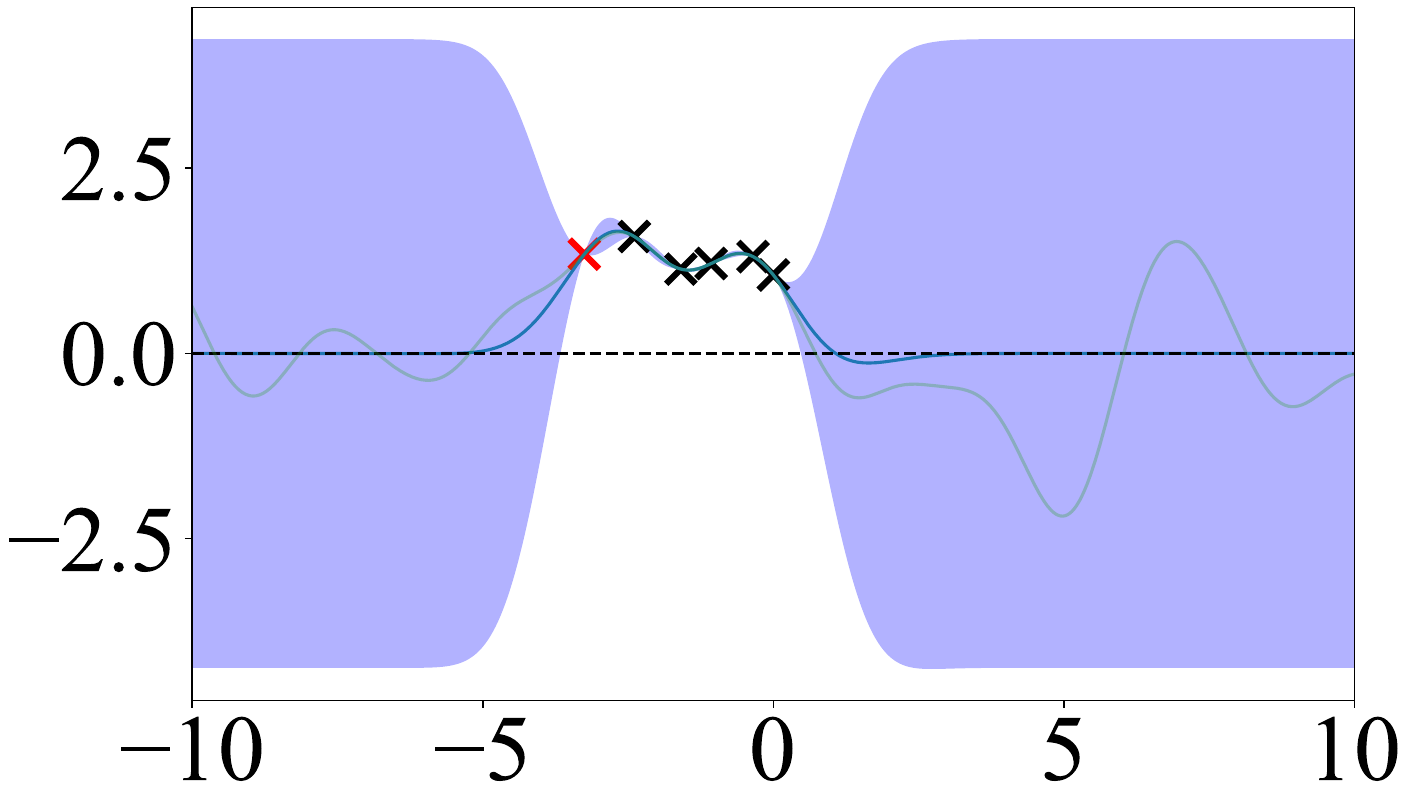}
         \caption{After 5 iterations.}
     \end{subfigure}
     \begin{subfigure}[b]{0.32\textwidth}
         \centering
         \includegraphics[width=\textwidth]{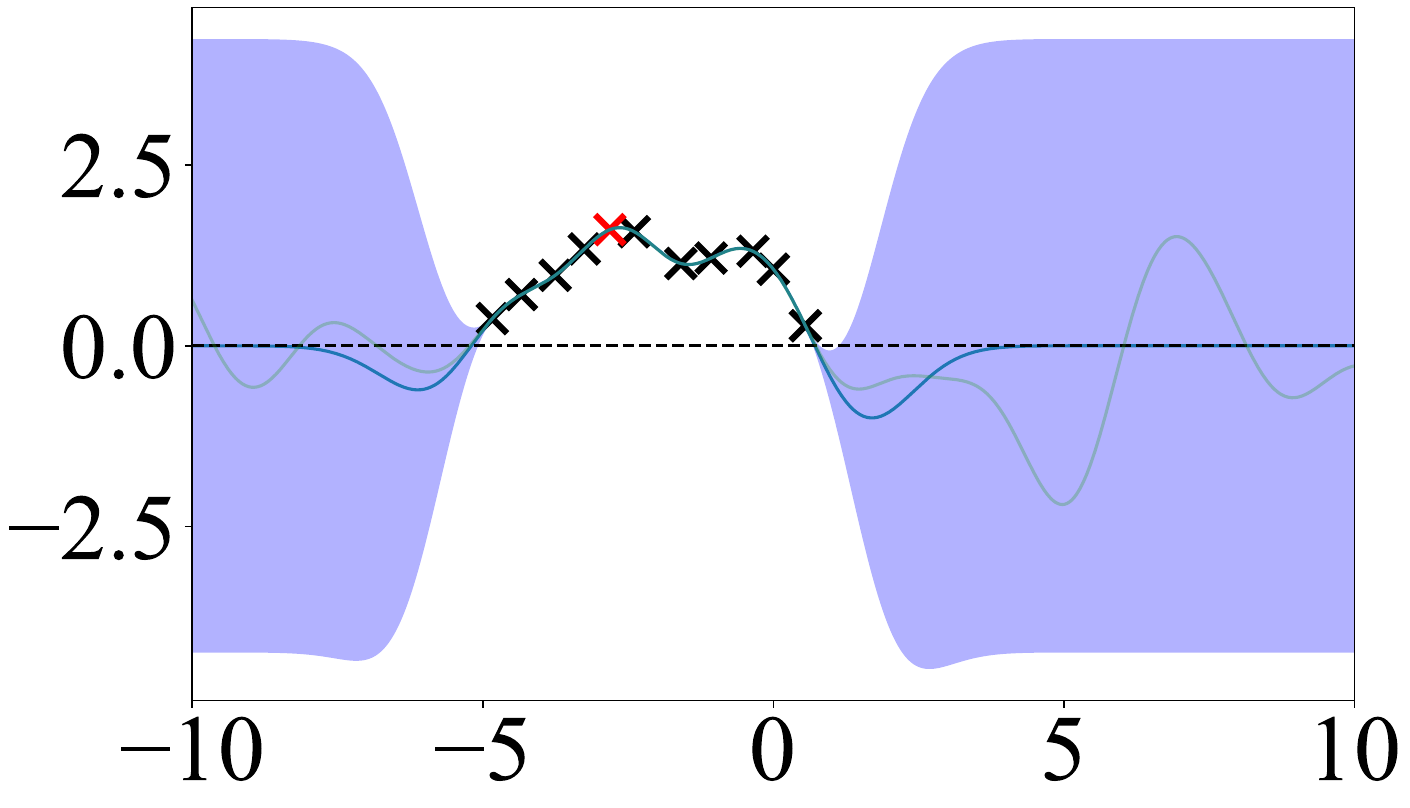}
         \caption{After 10 iterations.}
     \end{subfigure}
        \caption{1D visualization for the safe optimization process. The blue curve and purple shading represent the mean and confidence interval of the Gaussian process, respectively. The green curve represents the true value of the unknown function. Red markers indicate the prediction points acquired in the current iteration, and black markers show the previous prediction points. The black dashed line represents the safety threshold, which we set to 0. (a) - (c) are the results using the potential expander set $\mathcal{E}_n$, and (d) - (f) are the results using the set of safe boundary points $\mathcal{B}_n$.}
        \label{fig:1D_visualization}
\end{figure}

\begin{figure}[ht]
     \centering
     \begin{subfigure}[b]{0.31\textwidth}
         \centering
         \includegraphics[width=\textwidth]{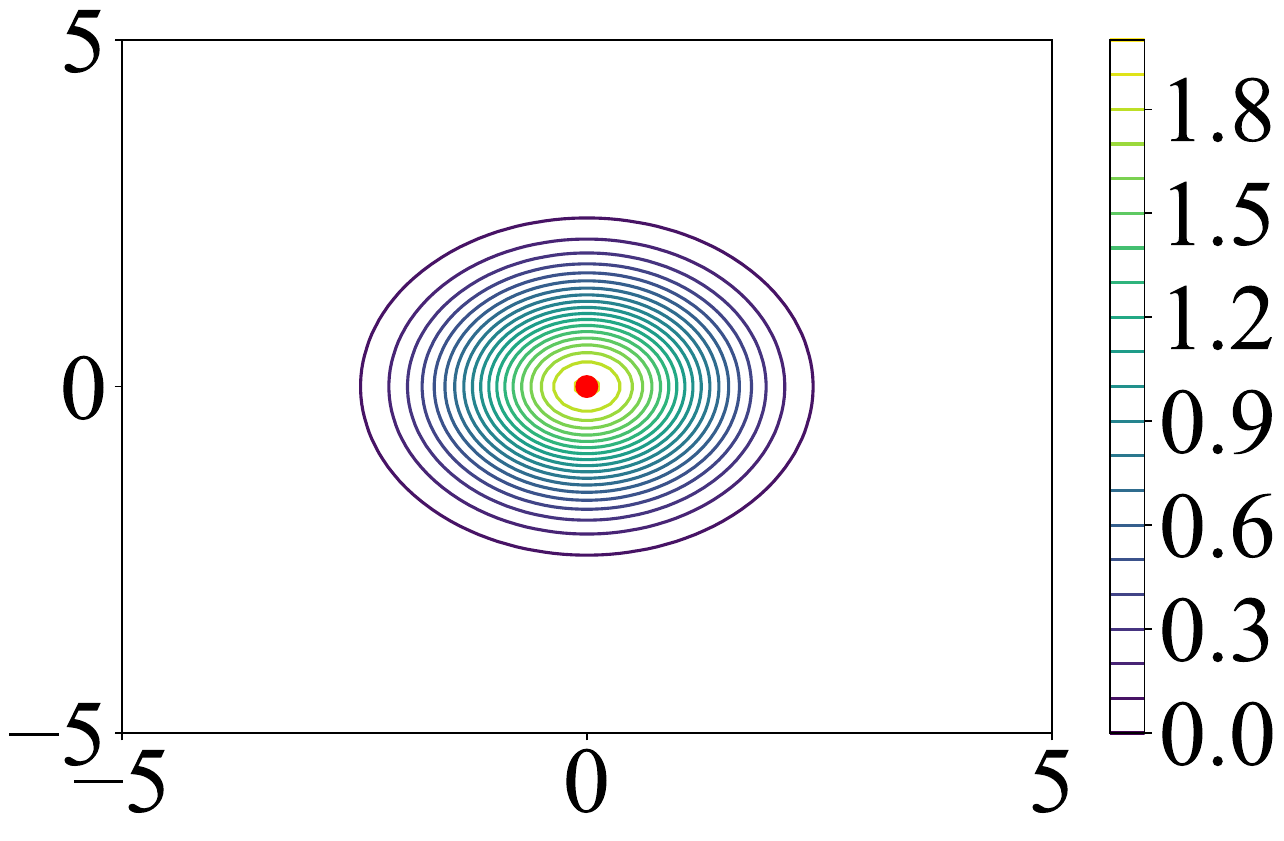}
         \caption{Initial safe prior.}
     \end{subfigure}
     \begin{subfigure}[b]{0.32\textwidth}
         \centering
         \includegraphics[width=\textwidth]{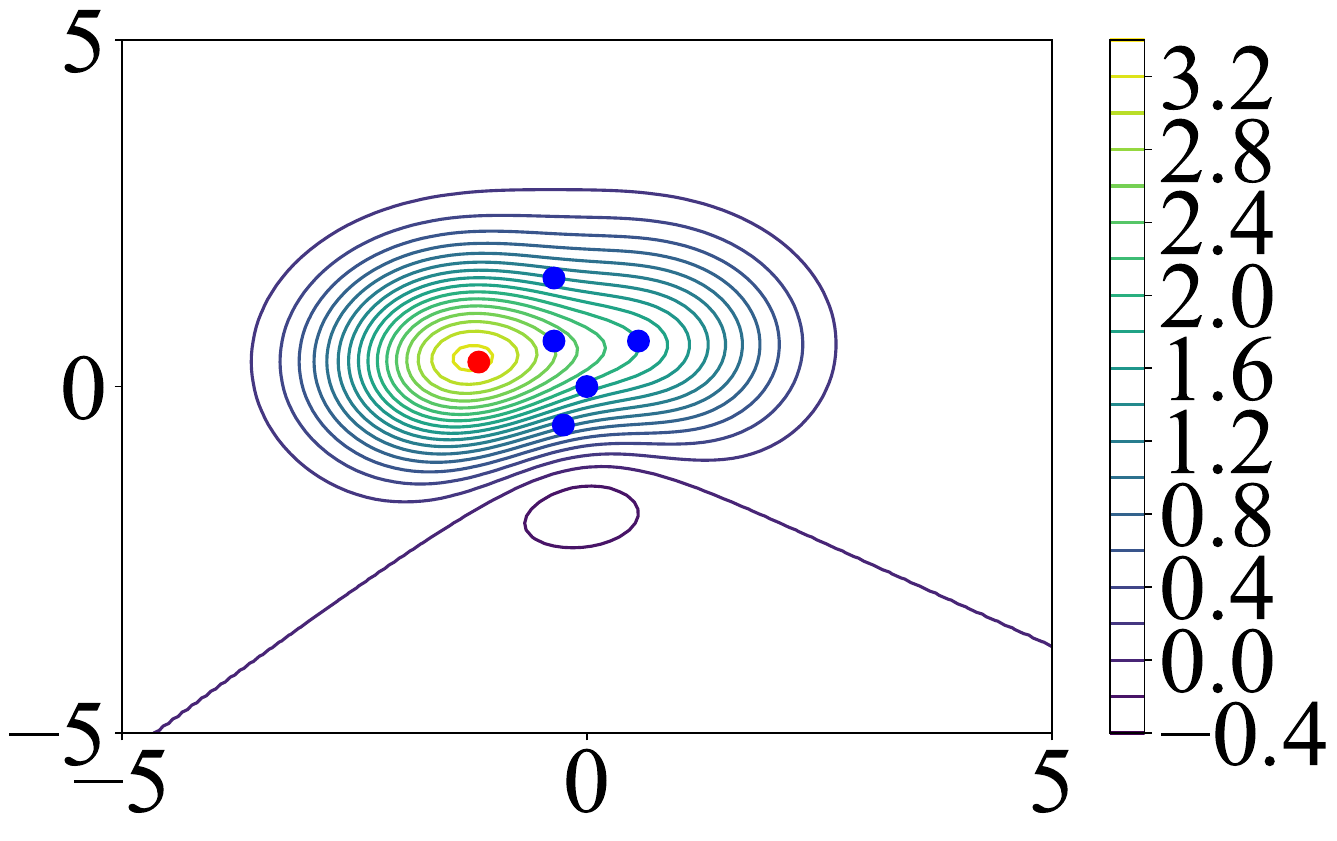}
         \caption{After 5 iterations.}
     \end{subfigure}
     \begin{subfigure}[b]{0.32\textwidth}
         \centering
         \includegraphics[width=\textwidth]{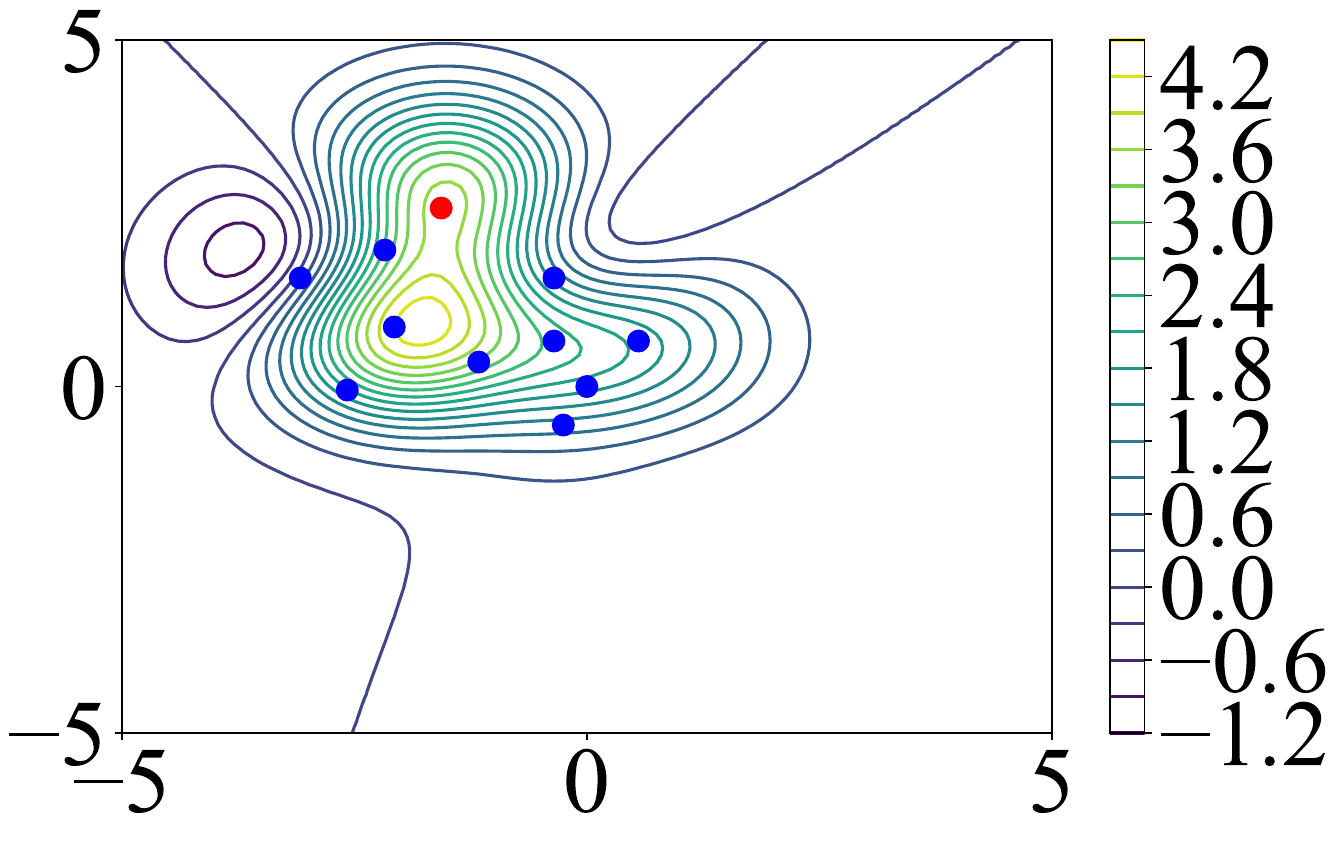}
         \caption{After 10 iterations.}
     \end{subfigure}
     \begin{subfigure}[b]{0.31\textwidth}
         \centering
         \includegraphics[width=\textwidth]{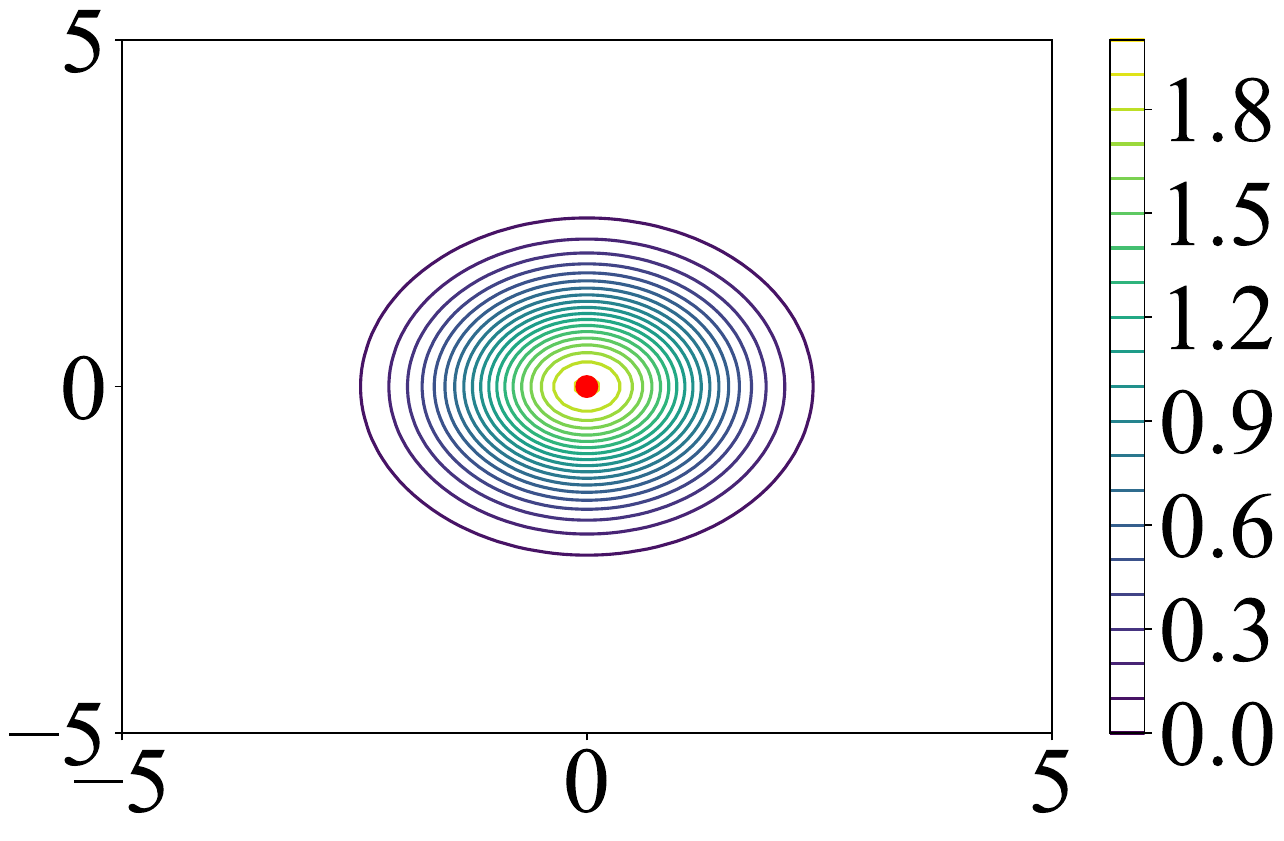}
         \caption{Initial safe prior.}
     \end{subfigure}
     \begin{subfigure}[b]{0.32\textwidth}
         \centering
         \includegraphics[width=\textwidth]{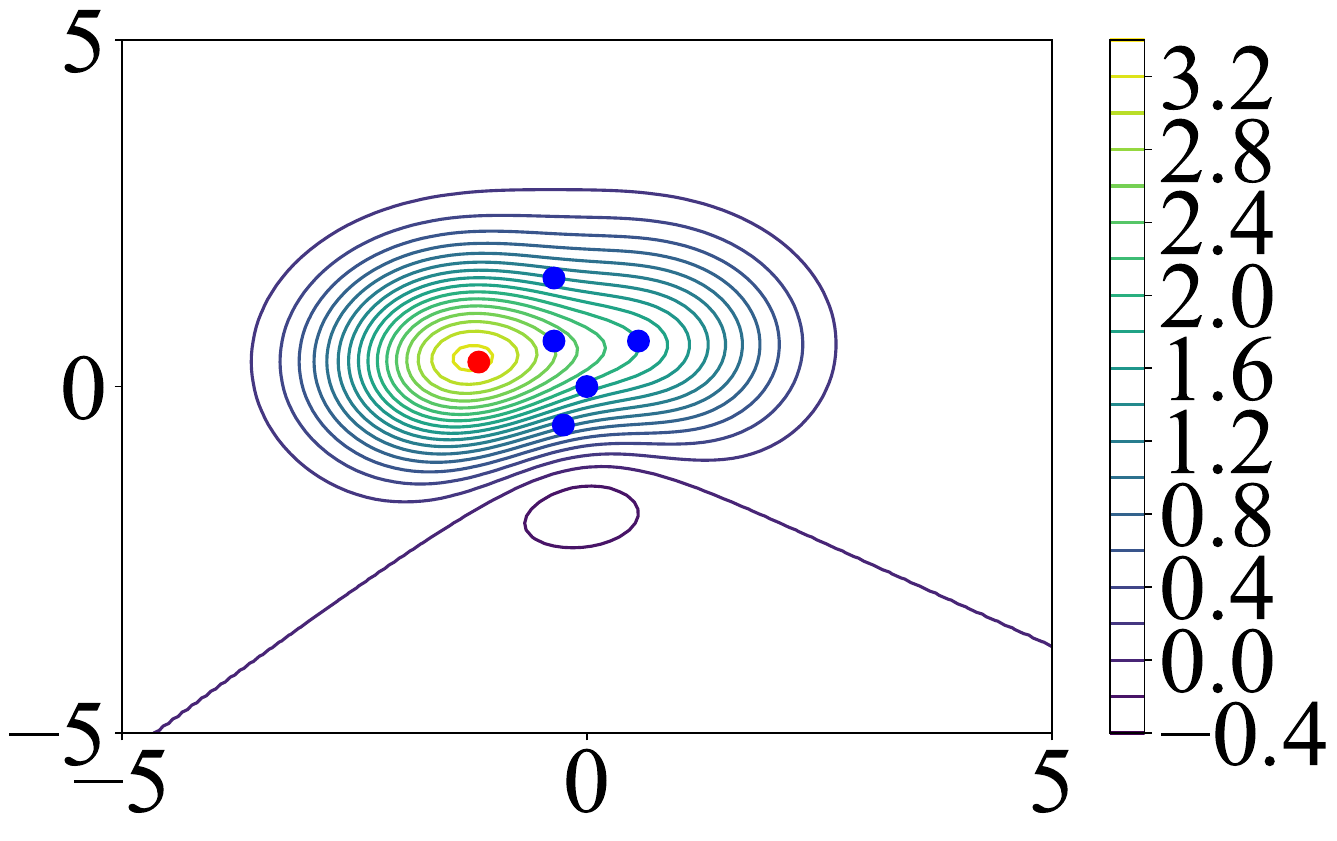}
         \caption{After 5 iterations.}
     \end{subfigure}
     \begin{subfigure}[b]{0.32\textwidth}
         \centering
         \includegraphics[width=\textwidth]{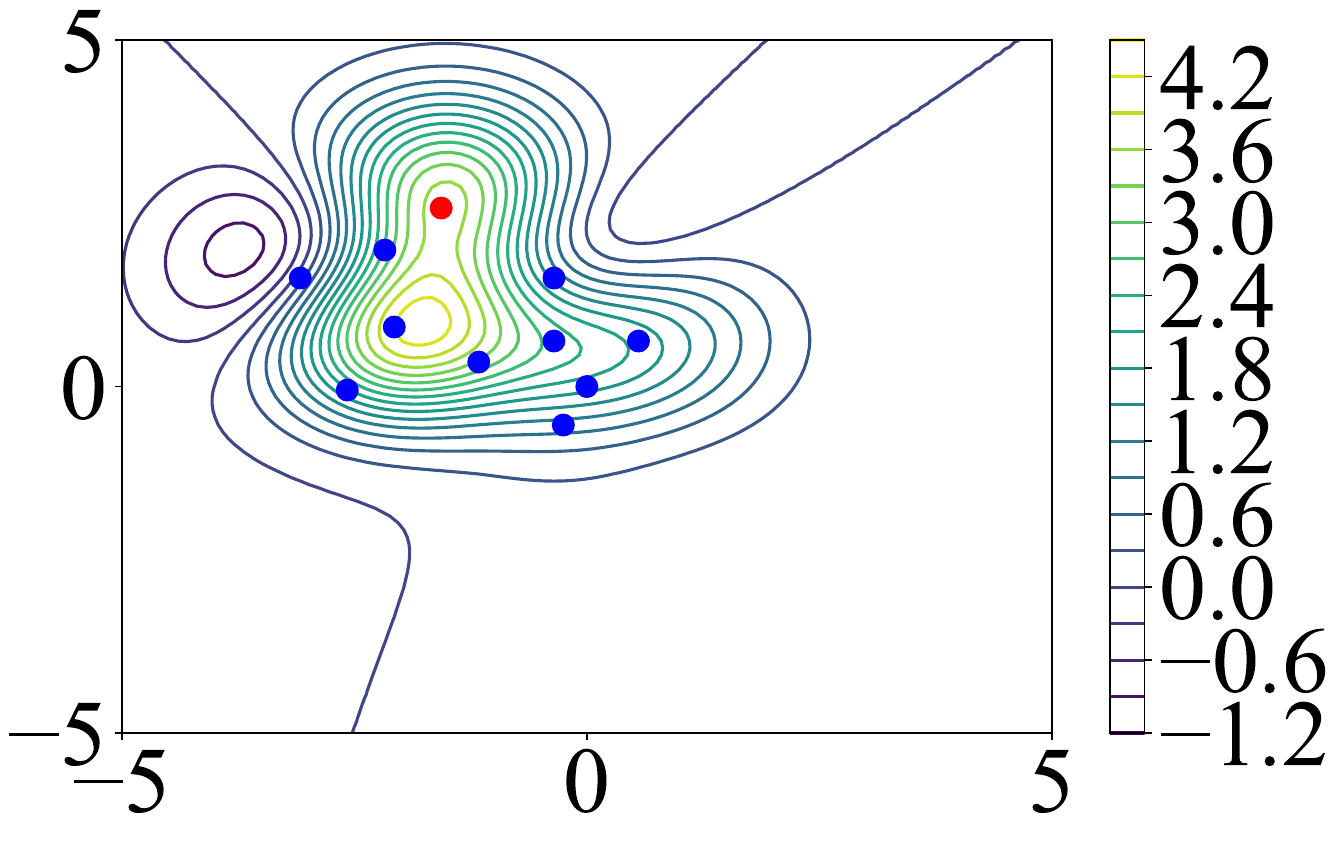}
         \caption{After 10 iterations.}
     \end{subfigure}
        \caption{2D visualization for safe exploration. The contour values represent the predicted values of the unknown function by the Gaussian process. Red markers indicate the prediction points acquired in the current iteration, while blue markers represent the previous prediction points. (a) - (c) are the results using the potential expander set $\mathcal{E}_n$, and (d) - (f) are the results using the set of safe boundary points $\mathcal{B}_n$.}
        \label{fig:2D_visualization}
\end{figure}

\paragraph{Visualization in 1D and 2D cases.} We show the effectiveness of simplifying the safe expansion stage visually in 1D and 2D. We use \textsc{SafeOpt} as the baseline algorithm and replace the potential expander set $\mathcal{E}_n$ used in \textsc{SafeOpt} with the proposed set of safe boundary points $\mathcal{B}_n$ as the comparison algorithm. Safe optimization was performed on 1D and 2D simulation functions with the safety threshold set to 0. As shown in Figure \ref{fig:1D_visualization} and \ref{fig:2D_visualization}, the comparison algorithm using $\mathcal{B}_n$ can acquire the same prediction points as \textsc{SafeOpt}.

\subsection{Proofs for Theoretical Results}

\subsubsection{Proof of Theorem \ref{thm4}}
\label{proof_thm4}

The proof uses a standard kernel-metric argument. For a positive definite kernel $k$, define
\begin{equation}
    d_k(a,z)^2=k(a,a)+k(z,z)-2k(a,z).
\end{equation}
For stationary kernels, $k(a,a)=k(z,z)=S$. After one noiseless observation at $z$, the latent posterior variance at $a$ is
\begin{equation}
    \sigma^2(a\mid z)=S-\frac{k(a,z)^2}{S}
    =d_k(a,z)^2-\frac{d_k(a,z)^4}{4S}
    \le d_k(a,z)^2 .
\end{equation}
With observation noise, or with a finite number of repeated measurements, let $\rho_\eta^2$ upper-bound the remaining posterior variance at an evaluated point. Then
\begin{equation}
    \sigma^2(a\mid z)\le d_k(a,z)^2+\rho_\eta^2 .
\end{equation}
Additional observations can only reduce the latent posterior variance, so the same upper bound holds once at least one sampled point $z$ lies sufficiently close to $a$.

For a first-order additive squared-exponential kernel
\begin{equation}
    k_{\mathrm{add}}(a,z)=\sum_{j=1}^D \sigma_f^2
    \exp\left[-\frac{(a_j-z_j)^2}{2\ell_j^2}\right],
\end{equation}
the associated kernel metric satisfies
\begin{align}
    d_k(a,z)^2
    &=2\sum_{j=1}^D \sigma_f^2\left(1-
    \exp\left[-\frac{(a_j-z_j)^2}{2\ell_j^2}\right]\right) \\
    &\le \sum_{j=1}^D \sigma_f^2\frac{(a_j-z_j)^2}{\ell_j^2},
\end{align}
where the last inequality uses $1-e^{-x}\le x$. If the expansion samples form a coordinate-wise cover such that every $a\in\mathcal{R}_\epsilon$ has a sampled point $z$ with $|a_j-z_j|\le r_j$ and
\begin{equation}
    r_j \le \frac{\ell_j\epsilon_\sigma}{\sigma_f\sqrt{D}},
\end{equation}
then $d_k(a,z)^2\le\epsilon_\sigma^2$ and therefore
\begin{equation}
    \sigma_{t^*-1}^{(i)}(a)^2\le \epsilon_\sigma^2+\rho_\eta^2
    =\left(\frac{\epsilon}{2\beta_{t^*}}\right)^2 .
\end{equation}
Thus $2\beta_{t^*}\sigma_{t^*-1}^{(i)}(a)\le\epsilon$ for all $a\in\mathcal{R}_\epsilon$ and all safety functions. Since the confidence event holds with probability at least $1-\delta$, the resulting safe set is $\epsilon$-accurate with the same probability.

A hyper-rectangle with side lengths $L_j$ can be covered by at most
\begin{equation}
    \prod_{j=1}^{D}\left\lceil \frac{L_j}{r_j}\right\rceil
    \le
    \prod_{j=1}^{D}\left\lceil \frac{\sigma_f\sqrt{D}L_j}{\ell_j\epsilon_\sigma}\right\rceil
\end{equation}
axis-aligned cells, which gives \Eqref{eq:expansion-budget}. For higher-order additive kernels, replace the first-order coordinate metric above by the kernel-metric upper bound in \Eqref{eq:kernel-metric-lipschitz}; the same covering argument then applies with a different constant.

\subsubsection{Proof of Theorem \ref{thm5}}
\label{proof_thm5}

Let the maximization stage run for $T$ evaluations on the fixed safely reachable region $\mathcal{R}_\epsilon$, and let $a_t$ be the point selected by the upper-confidence rule in \Eqref{eq:max-acquisition}. On the confidence event,
\begin{equation}
    J(a)\in[\mu_{t-1}(a)-\beta_t\sigma_{t-1}(a),
    \mu_{t-1}(a)+\beta_t\sigma_{t-1}(a)]
\end{equation}
for every $a\in\mathcal{R}_\epsilon$ and every $t$. Because $a_t$ maximizes the upper confidence bound,
\begin{align}
    J(a^*)-J(a_t)
    &\le u_t(a^*)-\ell_t(a_t) \\
    &\le u_t(a_t)-\ell_t(a_t) \\
    &=2\beta_t\sigma_{t-1}(a_t).
\end{align}
The best observed point $a_{\mathrm{rec}}$ has no larger simple regret than the best single iterate, hence
\begin{equation}
    J(a^*)-J(a_{\mathrm{rec}})
    \le \min_{1\le t\le T}2\beta_t\sigma_{t-1}(a_t)
    \le 2\beta_T\sqrt{\frac{1}{T}\sum_{t=1}^T\sigma_{t-1}^2(a_t)},
\end{equation}
where we used that $\beta_t$ is nondecreasing.

For GP posteriors with bounded kernel variance and the information gain definition used in GP-UCB analyses, the variance-sum lemma gives
\begin{equation}
    \sum_{t=1}^T\sigma_{t-1}^2(a_t)\le 2\gamma_T,
\end{equation}
up to the conventional noise-normalization constant, which is absorbed into the definition of $\gamma_T$ used here. Substitution yields
\begin{equation}
    J(a^*)-J(a_{\mathrm{rec}})
    \le 2\beta_T\sqrt{\frac{2\gamma_T}{T}} .
\end{equation}
Therefore the sufficient condition
\begin{equation}
    \frac{8\beta_{T^*}^2\gamma_{T^*}}{T^*}\le\zeta^2
\end{equation}
implies $J(a^*)-J(a_{\mathrm{rec}})\le\zeta$, proving Theorem~\ref{thm5}.


\section{Ablation Study for Simplifying the Safe Expansion Stage}
\label{sec:ablation}

\begin{figure}[H]
    \centering
    
    \includegraphics[width=0.9\textwidth]{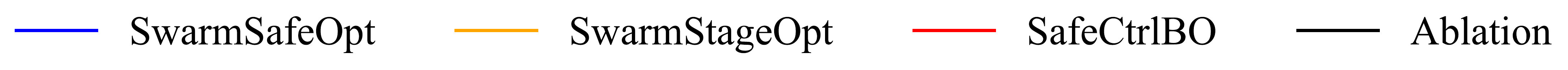}  
    \vspace{0.05cm}  
    
    \begin{tikzpicture}
    \setlength{\belowcaptionskip}{0pt}
        \node[anchor=south west,inner sep=0] (image) at (0,0) {
            \begin{tabular}{ccc}
                \hspace{-0.3cm}
                \begin{subfigure}[b]{0.32\textwidth}  
                    \includegraphics[width=\textwidth]{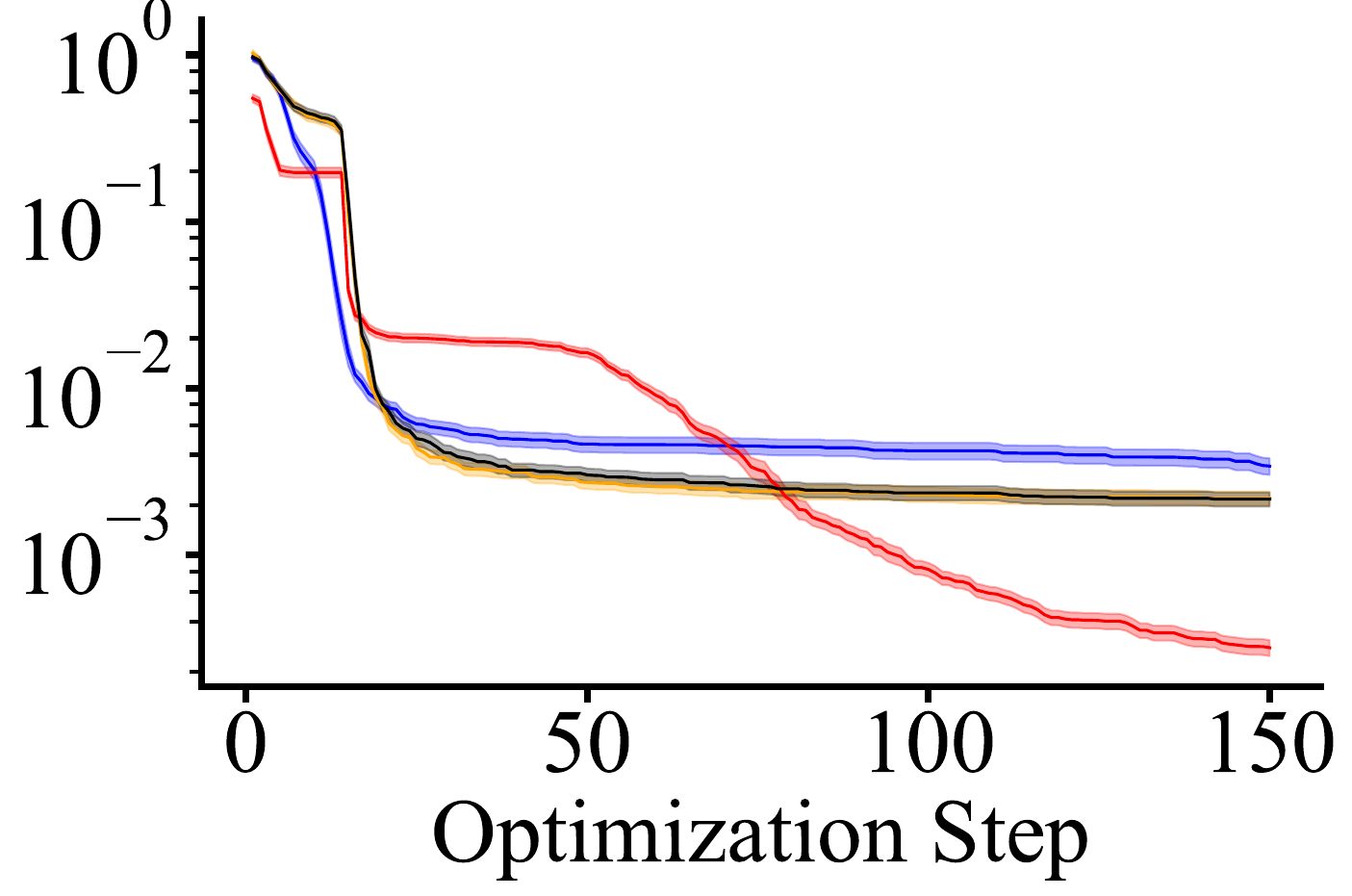}
                    \caption{Camelback2D}
                    \label{fig:camelback_ab}
                \end{subfigure} 
                \hspace{-0.21cm} 
                \begin{subfigure}[b]{0.32\textwidth}  
                \includegraphics[width=\textwidth]{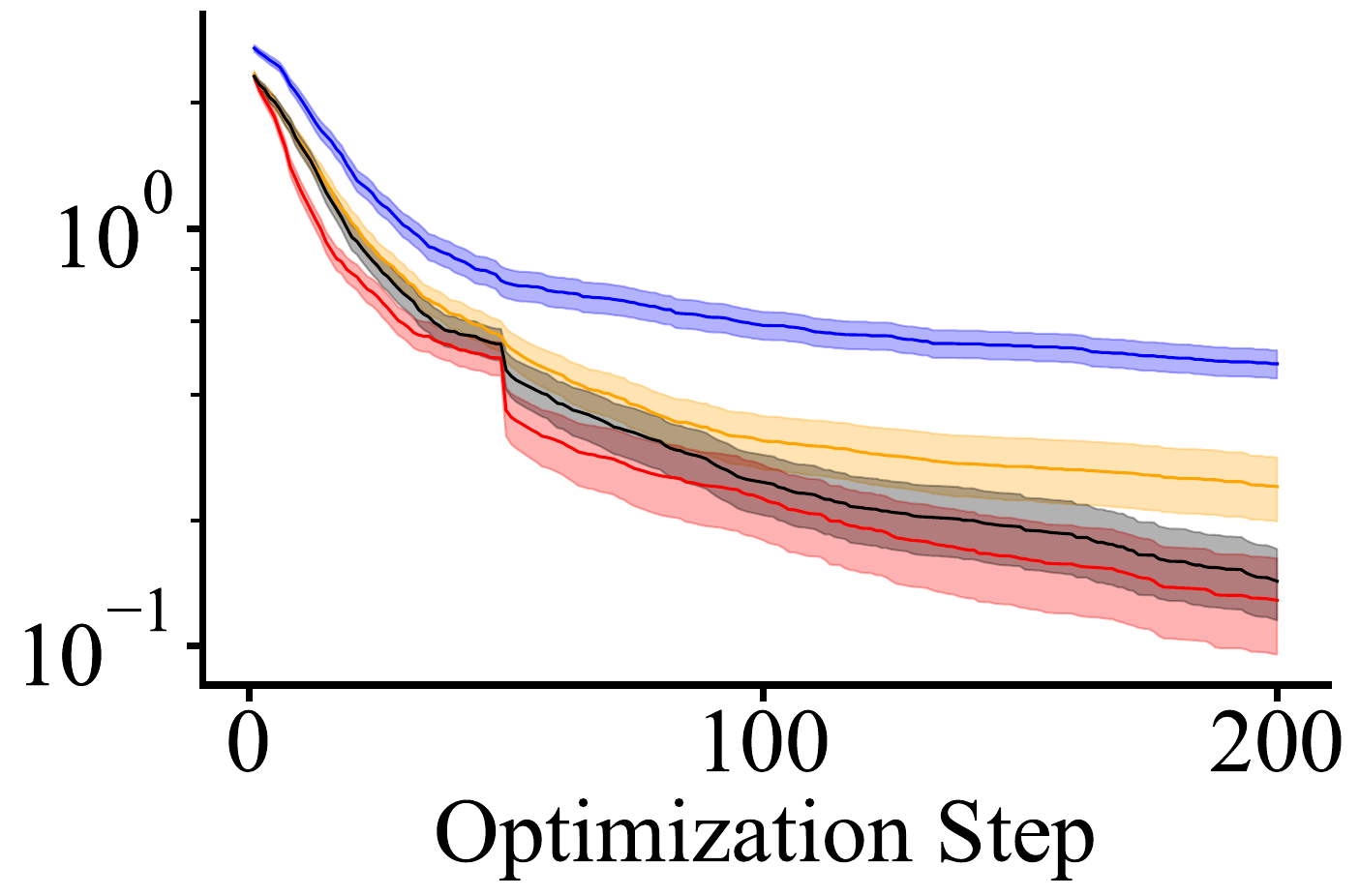}
                    \caption{Hartmann6D}
                    \label{fig:hartmann_ab}
                \end{subfigure} 
                \hspace{-0.21cm}
                \begin{subfigure}[b]{0.33\textwidth}  

                    \includegraphics[width=\textwidth]{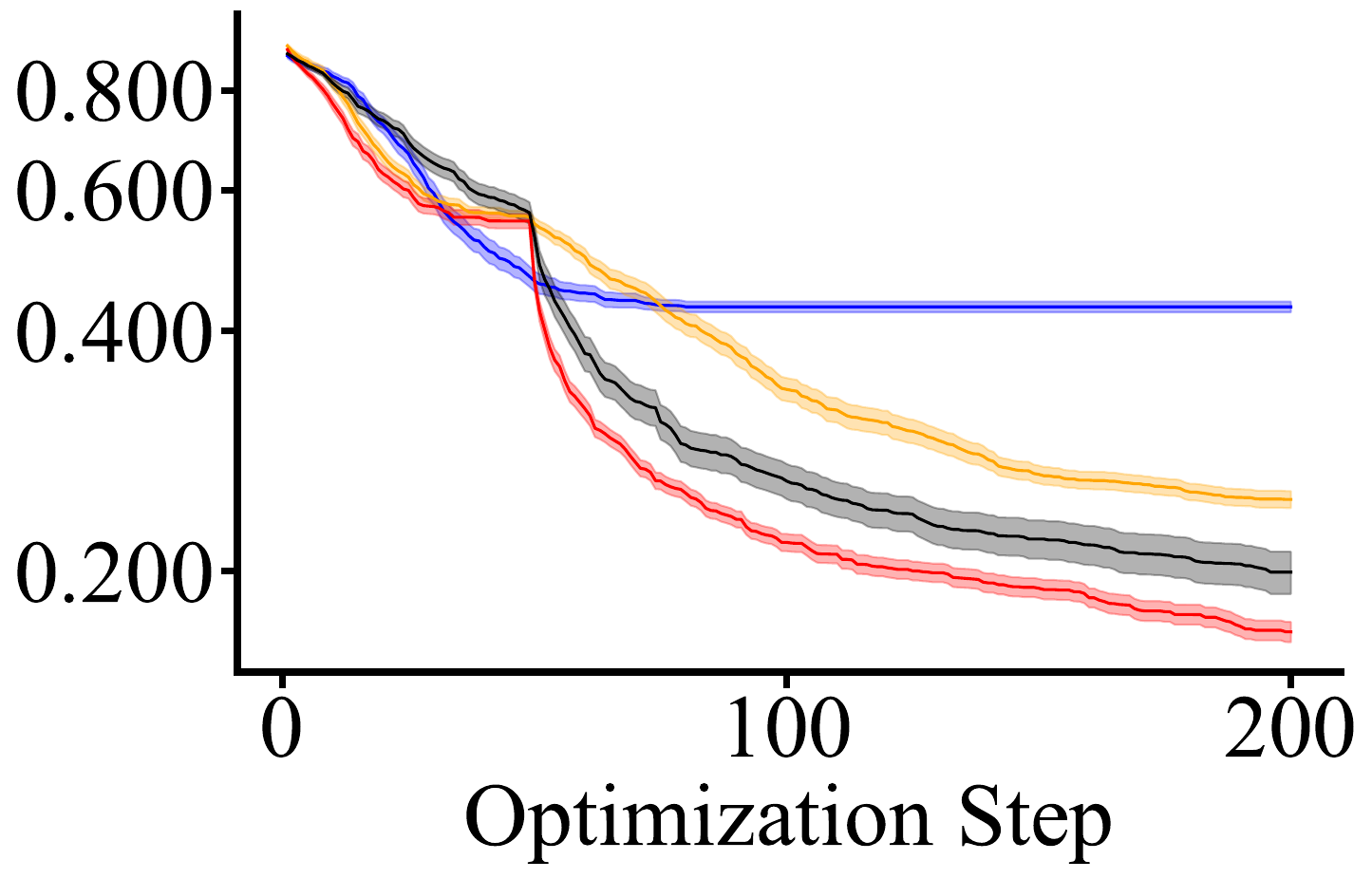}
                    \caption{Gaussian10D}
                    \label{fig:gaussian_ab}
                \end{subfigure} \\
            \end{tabular}
        };
        \node[rotate=90, anchor=south] at ([yshift=0.7cm,font=\tiny]image.west) {\small{Simple Regret}};
    \end{tikzpicture}
    \caption{Ablation study using the synthetic benchmark functions.}
    \label{fig:synthetic simulations ablation}
\end{figure}

In \textsc{SafeCtrlBO}, besides replacing Gaussian kernels with additive kernels, we also simplified the potential expander set $\mathcal{E}_n$ to the set of safe boundary points $\mathcal{B}_n$. In this section, we test whether the simplified safe exploration process results in any performance loss compared to previous safe BO methods.

Figure \ref{fig:synthetic simulations ablation} shows the optimization results of different algorithms on the three benchmark functions, with the black curve representing the Ablation algorithm, which uses $\mathcal{B}_n$ for safe exploration but does not employ additive kernels. Therefore, the only difference between the Ablation algorithm and \textsc{SwarmStageOpt} lies in the iteration strategy for safe exploration. As shown in Figure \ref{fig:synthetic simulations ablation}, the Ablation algorithm’s results on all three benchmark functions are not inferior to those of \textsc{SwarmSafeOpt} and \textsc{SwarmStageOpt}, suggesting that using the simplified safe exploration process does not lead to performance losses compared to using the potential expander set $\mathcal{E}_n$.

\end{document}